\documentclass{llncs}

\usepackage[utf8]{inputenc} 
\usepackage[T1]{fontenc}    
\usepackage{hyperref}       
\usepackage{url}            
\usepackage{booktabs}       
\usepackage{amssymb}
\usepackage{nicefrac}       
\usepackage{microtype}      
\usepackage[usenames,dvipsnames]{xcolor} 
\usepackage{amsmath} 
\usepackage{stmaryrd} 
\usepackage{adjustbox}
\usepackage{sidecap} 
\usepackage{float}
\usepackage[mathscr]{euscript}
\usepackage{footnote} 
\makesavenoteenv{tabular}
\newtheorem{observation}{Observation}

\usepackage[noend]{algpseudocode}
\usepackage{algorithm}
\algblock{ParFor}{EndParFor}
\algnewcommand\algorithmicparfor{\textbf{parfor}}
\algnewcommand\algorithmicpardo{\textbf{do}}
\algnewcommand\algorithmicendparfor{\textbf{end\ parfor}}
\algrenewtext{ParFor}[1]{\algorithmicparfor\ #1\ \algorithmicpardo}
\algrenewtext{EndParFor}{\algorithmicendparfor}
\algnewcommand\algorithmicinput{\textbf{Input:}}
\algnewcommand\algorithmicoutput{\textbf{Output:}}
\algnewcommand\Input{\item[\algorithmicinput]}%
\algnewcommand\Output{\item[\algorithmicoutput]}%

\usepackage{tablefootnote}
\usepackage{changepage}   

\usepackage{xcolor}

\usepackage[small]{subfigure}
\usepackage{stackengine}

\usepackage{color, colortbl}
\definecolor{Gray}{gray}{0.9}

\begin{document}
%
%
\title{Reachability Analysis of a General Class of Neural Ordinary Differential Equations}
\author{%
Diego Manzanas Lopez\inst{1} \and 
Patrick Musau\inst{1} \and 
Nathaniel P. Hamilton\inst{1} \and 
Taylor T. Johnson\inst{1}
}
\institute{Vanderbilt University,  Nashville, TN 37212, USA \\
\email{$\{$diego.manzanas.lopez, taylor.johnson$\}$@vanderbilt.edu}
}
\pagestyle{plain}
%
%
\maketitle

\begin{abstract}
Continuous deep learning models, referred to as Neural Ordinary Differential Equations (Neural ODEs), have received considerable attention over the last several years. Despite their burgeoning impact, there is a lack of formal analysis techniques for these systems. In this paper, we consider a general class of neural ODEs 
with varying architectures and layers, and  introduce a novel reachability framework that allows for the formal analysis of their behavior. The methods developed for the reachability analysis of neural ODEs are implemented in a new tool called NNVODE. Specifically, our work extends an existing neural network verification tool to support neural ODEs. We demonstrate the capabilities and efficacy of our methods through the analysis of a set of benchmarks that include neural ODEs used for classification, and in control and dynamical systems, including an evaluation of the efficacy and capabilities of our approach with respect to existing software tools within the continuous-time systems reachability literature, when it is possible to do so. \let\thefootnote\relax\footnotetext{Extended version of the paper to appear at FORMATS 2022 \cite{manzanas2022formats}.}


\end{abstract}

\section{Introduction}
Neural Ordinary Differential Equations (ODEs) were first introduced in 2018, as a radical new neural network design that boasted better memory efficiency, and an ability to deal with irregularly sampled data \cite{hao_2020}. The idea behind this family of deep learning models is that instead of specifying a discrete sequence of hidden layers, we instead parameterize the derivative of the hidden states using a neural network \cite{chen2018neural}. The output of the network can then be computed using a differential equation solver \cite{chen2018neural}. This work has spurred a whole range of follow-up work, and since 2018 several variants have been proposed, such as augmented neural ODEs (ANODEs) and their ensuing variants \cite{norcliffe2020sonode,dupont2019neurips,amir2019anode}. These variants provide a more expressive formalism by augmenting the state space of neural ODEs to allow for the flow of state trajectories to cross. This crossing, prohibited in the original framework, allows for the learning of more complex functions that were prohibited by the original neural ODE formulation \cite{dupont2019neurips}.

Due to the potential that neural networks boast in revolutionizing the development of intelligent systems in numerous domains, the last several years have witnessed a significant amount of work towards the formal analysis of these models.
The first set of approaches that were developed considered the formal verification of neural networks (NN), using a variety of techniques including reachability methods \cite{tran2020cavtool,wang2018reluval,bak2021nnenum,ruan2018ijcai}, and SAT techniques \cite{katz2017reluplex,katz2019marabou,ehlers2017atva}. Thereafter, many researchers proposed novel formal method approaches for neural network control systems (NNCS), where the majority of methods utilized a combination of NN and hybrid system verification techniques \cite{fan2020reachNN*,Huang2019reachNN,tran2019emsoft,Ivanov2019verisig,ivanov2021verisig20,bogomolov2019Juliareach}. Building on this work, a natural outgrowth is extending these approaches to analyze and verify neural ODEs, 
and some recent studies have considered the analysis of formal properties of neural ODEs. One such study aims to improve the understanding of the inner operation of these networks by analyzing and experimenting with multiple neural ODE architectures on different benchmarks \cite{massaroli2020neurips}. Some studies have considered analyses of the robustness of neural ODEs such as \cite{carrara2019wifs} and \cite{yan2020robustness}, which evaluate the robustness of image classification neural ODEs and compare the efficacy of this class of network against other more traditional image classifier architectures. The first reachability technique targeted for neural ODEs presented a theoretical regime for verifying neural ODEs using Stochastic Lagrangian Reachability (SLR) \cite{gruenbacher2021verification}. This method is an abstraction-based technique that computes confidence intervals for the calculated reachable set with probabilistic guarantees. In a follow-up work, these methods were improved and implemented in a tool called Gotube \cite{gruenbacher2021gotube}, which is able to compute reach sets for longer time horizons than most state-of-the-art reachability tools. However, these methods only provide stochastic bounds on the reach sets, so there are no formal guarantees on the derived results. 


To the best of our knowledge, this paper presents the first deterministic verification framework for a general class of neural ODEs with multiple continuous-time and discrete-time layers. In this work, we present our verification framework NNVODE that makes use of deterministic reachability approaches for the analysis of neural ODEs. 
Our methods are evaluated on a set of benchmarks with different architectures and conditions in the area of dynamical systems, control systems, and image classification. We also compare our results against three state-of-the-art verification tools when possible.
In summary, the contributions of this paper are:
\begin{itemize}
    \item We introduce a general class of neural ODEs that allows for the combination of multiple continuous-time and discrete-time layers.
    \item We develop NNVODE, an extension of NNV \cite{tran2020cavtool}, to formally analyze a general class of neural ODEs using sound and deterministic reachable methods.
    \item We run an extensive evaluation on a collection of benchmarks within the context of time-series analysis, control systems, and image classification. We compare the results to Flow*, GoTube, and JuliaReach for neural ODE architectures where this is possible.
\end{itemize}

\section{Background and Problem Formulation}

Neural ODEs emerged as a continuous-depth variant of neural networks becoming a special case of ordinary differential equations (ODEs), where the derivatives are defined by a neural network, and can be expressed as:
\begin{equation}
\label{eq:omegalayer_func}
    \dot{z} = g(z), \quad \quad \quad  z(t_0) = z_0,
\end{equation}
where the dynamics of the function $g:\mathbb{R}^j\rightarrow{}\mathbb{R}^p$ are represented as a neural network, and initial state $z_0 \in \mathbb{R}^{j}$, where $j$ corresponds to the dimensionality of the states $z$. A \emph{neural network} (NN) is defined to be a collection of consecutively connected $\mathscr{NN}$-Layers as described in Definition \ref{def:nnLayer}.

\begin{definition}[$\mathscr{NN}$-Layer]
\label{def:nnLayer}
    A \textbf{$\mathscr{NN}$-Layer} is a function h: $\mathbb{R}^{j} \rightarrow{} \mathbb{R}^{p}$, with input x $\in$ $\mathbb{R}^{j}$, output y $\in$ $\mathbb{R}^{p}$
    defined as follows
    \begin{gather}
        y = h(x)
    \end{gather}
\end{definition}
where the function $h$ is determined by parameters $\theta$, typically defined as a tuple $\theta$ = $\langle$$\sigma$,\textbf{W},\textbf{b}$\rangle$ for fully-connected layers, where \textbf{W} $\in$ $\mathbb{R}^{j\times p}$, \textbf{b} $\in$ $\mathbb{R}^{p}$, and activation function $\sigma$: $\mathbb{R}^{j} \rightarrow{} \mathbb{R}^{p}$, thus the fully-connected $\mathscr{NN}$-Layer is described as
\begin{gather}\label{eq:nnlayer}
    y = h(x) = \sigma(\textbf{W}(x) + \textbf{b}). 
\end{gather}

\begin{figure}[!ht]
  \centering
    \includegraphics[trim={300 170 300 230}, clip, width=0.55\columnwidth]{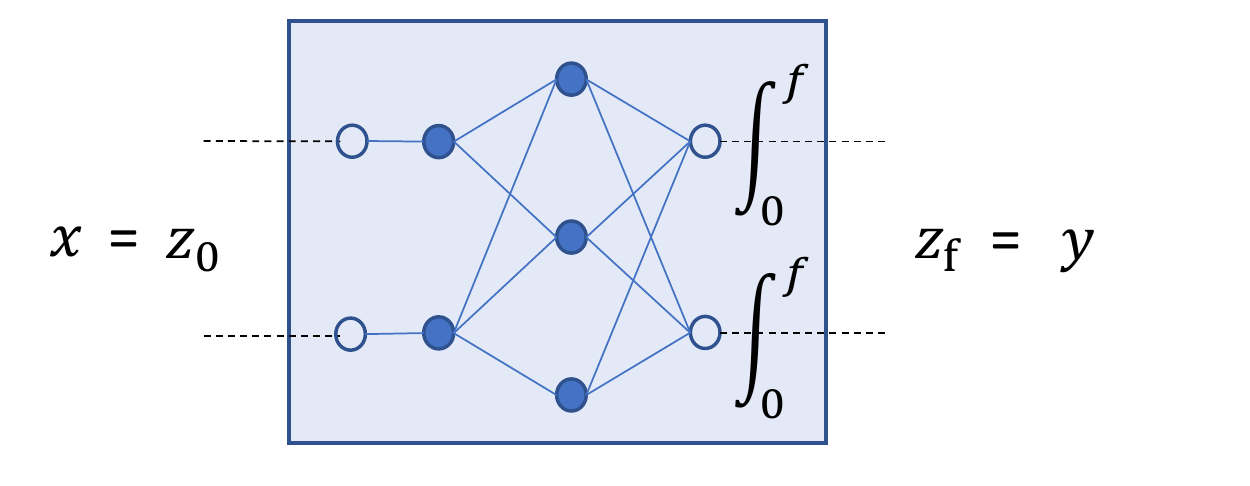}
   \caption{Illustration of an example of NODE, as defined in \eqref{eq:omegalayer_func} and ~\eqref{eq:omegalayer_layer}, and Definition \ref{def:omegaLayer}.}
  \label{fig:omegalayer}
\end{figure}

However, for other layers such as convolutional-type $\mathscr{NN}$-Layers, $\theta$ may include parameters like the filter size, padding, or dilation factor and the function $h$ in ~\eqref{eq:nnlayer} may not necessarily apply. For a formal definition and description of the reachability analysis for each of these layers that are integrated within NNVODE, we refer the reader to Section 4 of \cite{tran2020cav}. For the Neural ODEs (NODEs),
we assume that $g$ is Lipschitz continuous, which guarantees that the solution of $\dot{z}=g(z)$ exists. This assumption allows us to model $g$ with $m$ layers of continuously differentiable activation functions $\sigma_k$ for $k\in\{1,2,...,m\}$ such as sigmoid, tanh, and exponential activation functions. Described in ~\eqref{eq:omegalayer_func} is the notion of a NODE as introduced in \cite{chen2018neural} and an example is illustrated in Figure \ref{fig:omegalayer}. 


\begin{definition}[NODE]
\label{def:omegaLayer}
    A \textbf{NODE} is a function $\dot{z} = g(z)$ with $m$ fully-connected $\mathscr{NN}$-Layers, and it is defined as follows
    \begin{gather}\label{eq:omegalayer_layer}
        \dot{z} = g(z) = h_m(h_{m-1}(...h_1(z))),
    \end{gather}
    where $g : \mathbb{R}^j \rightarrow{} \mathbb{R}^p$, $\sigma_k : \mathbb{R}^{j_k} \rightarrow{} \mathbb{R}^{p_k}$, $\textbf{W}_k \in \mathbb{R}^{j_k \times p_k}$, and $\textbf{b}_k \in \mathbb{R}^{p_k}$. For each layer $k$=$\{1,2,\dots,m\}$, we describe the function $h_k$ as in ~\eqref{eq:nnlayer}.
\end{definition}

%
An example of NODE with m~=~2 hidden layers, 2 inputs and 2 outputs is depicted in Figure~\ref{fig:omegalayer}.

\subsection{General Neural ODE}
The general class of neural ODEs (GNODEs) considered in this work is more complex than previously analyzed neural ODEs as it may be comprised of two types of layer: NODEs and $\mathscr{NN}$-Layers.
We introduce a more general framework where multiple NODEs can make up part of the overall architecture along with other $\mathscr{NN}$-Layers, as described in Definition \ref{def:neuralODE}. This is the reachability problem subject of evaluation in this work.

\begin{definition}[GNODE] \label{def:neuralODE}
A \textbf{GNODE} $~\mathcal{F}$ is any sequence of consecutively connected $N$ layers $\mathcal{L}_k$ for $k \in \{1, \dots N\}$ with $N_O$ NODEs and $N_{D}$ $\mathscr{NN}$-Layers, that meets the conditions $1 \leq N_O \leq N$, $0 \leq N_{D} < N$, and $N_D$ + $N_O$ = $N$.
\end{definition}

With the above definition, we formulate a theorem for the restricted class of neural ODEs (NODE) that we use to compare our methods against existing techniques. 

\begin{observation}[Special case 1: NODE]\label{obs:omegalayer}
Let $\mathcal{F}$ be a GNODE with $N$ layers. If $N = 1$, $N_O = 1$ and $N_D = 0$, then $\mathcal{F}$ is equivalent to an ODE whose continuous-time dynamics are defined as a neural network, and we refer to it as a \textbf{NODE} (as in Definition~\ref{def:omegaLayer}).
\end{observation}

\begin{figure}[!ht]
  \centering
    \includegraphics[trim=0 0 0 0, clip, width=0.95\columnwidth]{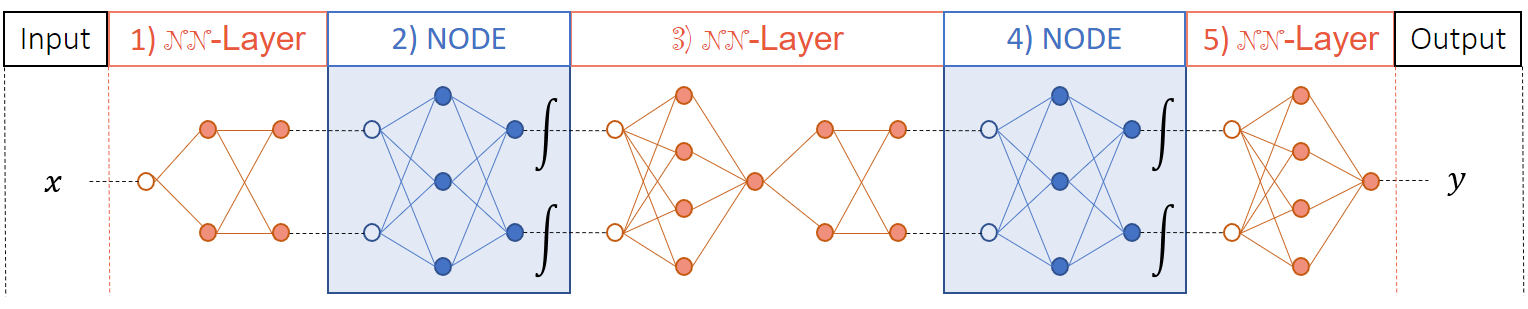}
    \vspace{-1em}
  \caption{Example of a GNODE. The filled circles represent weighted neurons in each layer, non-filled neurons represent the inputs of each layer-type segment, shown for visualization purposes. }
  \label{fig:gnode}
\end{figure}

In Figure \ref{fig:gnode}, an example of a GNODE is shown, which has 1 input (\emph{x}), 1 output (\emph{y}), $N_O$ = 2 NODEs, and $N_D$ = 5 $\mathscr{NN}$-Layers, with its 5 numbered segments described as: 1) The first segment has a $\mathscr{NN}$-Layer, with one hidden layer of 2 neurons and an output layer of 2 neurons, 2) a NODE with one hidden layer of 3 neurons and an output layer of 2 neurons, 3) $\mathscr{NN}$-Layer with 3 hidden layers of 4,1, and 2 neurons respectively, and an output layer of 2 neurons, 4) NODE with a hidden layer of 3 neurons and output layer of 2 neurons, and 5) $\mathscr{NN}$-Layer with a hidden layer of 4 neurons and and output layer with 1 neurons. 

Given this GNODE architecture, we are capable of encoding the originally proposed neural ODE \cite{chen2018neural}, as well as several of its model improvements including higher-order neural ODEs like the second-order neural ODE (SONODE) \cite{norcliffe2020sonode}, augmented neural ODEs (ANODEs) \cite{dupont2019neurips}, and input-layer augmented neural ODEs (ILNODEs). 
%
This general class of neural ODEs from Definition~\ref{def:neuralODE} is applicable to many applications including image classification, time series prediction, dynamical system modeling and normalizing flows.

\begin{observation}[Special case 2 - NNCS]\label{obs:nncs}
We can consider the NNCS as a special case of the \emph{GNODE}, as the NNCS models also consist of $\mathscr{NN}$-Layers and NODE, under the assumption that the plant is described as an NODE. NNCS in the general architecture have $\mathscr{NN}$-Layers followed by an ODE or NODE, which connects back to the $\mathscr{NN}$-Layers in a feedback loop manner for $cp$ control steps. 
Thus, by unrolling the NNCS $cp$ times, we consecutively connect the $\mathscr{NN}$-Layers with the NODE, creating a \emph{GNODE}.  
\end{observation}

\subsection{NODE: Applications} \label{sec:omegalayerapps}
There are two application modes of a NODE, model reduction and dynamical systems. On one hand, we can use it as a substitute for multiple similar layers to reduce the depth and overall size of a model \cite{chen2018neural}. In this context, we treat the NODE as an input-output mapping function, and set the integration time $t_f$ to a fixed number during training, which will be then used during simulation or reachability. Typically, this value is set to 1. On the other hand, we can make use of NODEs to capture the behavior of time-series data or dynamical systems. In this sense, $t_f$ is not fixed, and it will be determined by the user/application. In summary, we can use NODEs for: 1) time-series or dynamical system modeling, and 2) model reduction. For time-series, $t_f$ is variable, user or application dependent. For model reduction, $t_f$ is a parameter of the model fixed before training. Only the output value at time = $t_f$ is used, while for the time-series models, we are interested in the interval [0,$t_f$]. 

\subsection{Reachability Analysis} \label{sec:reachAnalysis}

The main focus of this manuscript is to introduce a framework for the reachability analysis of GNODEs. This framework
%
combines set representations and methods from Neural Network (NN), Convolutional Neural Network (CNN) and hybrid systems reachability analysis. Similar to neural network reachability in NNV \cite{tran2020cavtool}, we consider the set propagation through each layer as well as the set conversion between layers for all reachability methods for the supported layers. Hence, the reachability problem of GNODE is defined as follows:

\begin{definition}[Reachable Set of a GNODE]
    \label{def:NODEreach}
    Let $\mathcal{F}$: $\mathbb{R}^j \rightarrow{} \mathbb{R}^p$ be a \textbf{GNODE} with $N$ layers. The output reachable set $\mathcal{R}_{N}$ of a \textbf{GNODE} with input set $\mathcal{R}_0$ is defined as:
    \begin{equation}
    \begin{split}
        \mathcal{R}_{1} &\triangleq \{y_1~|~y_1 = f_1(y_0), ~ y_0 \in \mathcal{R}_0\}, \\
      \mathcal{R}_{2} &\triangleq \{y_2~|~y_2 = f_2(y_1), ~ y_1 \in \mathcal{R}_{1}\}, \\
     &\vdots \\
    \mathcal{R}_{N} &\triangleq \{y_N ~|~ y_N = f_N(y_{N-1}), ~y_{N-1} \in \mathcal{R}_{N-1} \}, \\
    \end{split}
    \end{equation}
where $f_i$: $\mathbb{R}^{j_i} \rightarrow{} \mathbb{R}^{p_i}$ is the function $f$ of layer $i$, where $f$ is either a NODE (g) or a $\mathscr{NN}$-Layer (h).
\end{definition}

The proposed solution to this problem is sound and incomplete, in other words, an over-approximation of the reachable set. This means that, given a set of inputs, we compute an output set of which, given any point in the input set, we simulate the GNODE and the output point is always contained within the reachable output set (sound). However, it is incomplete because the opposite is not true; there may be points in the output reachable set that cannot be traced back to be within the bounds of the input set.

\begin{definition}[Soundness] \label{def:sound}
Let  $\mathcal{F}$: $\mathbb{R}^j$ $\rightarrow{\mathbb{R}^p}$ be a GNODE with an input set $\mathcal{R}_0$ and output reachable set $\mathcal{R}_f$. The computed $\mathcal{R}_f$ given $\mathcal{F}$ and $\mathcal{R}_0$ is \textbf{sound} iff $\forall$$x$ $\in$ $\mathcal{R}_0$, $|$ $y$ = $\mathcal{F}$($x$), $y$ $\in$ $\mathcal{R}_f$. 
\end{definition}


\begin{definition}[Completeness] \label{def:complete}
Let  $\mathcal{F}$: $\mathbb{R}^j$ $\rightarrow{\mathbb{R}^p}$ be a GNODE with an input set $\mathcal{R}_0$ and output reachable set $\mathcal{R}_f$. The computed $\mathcal{R}_f$ given $\mathcal{F}$ and $\mathcal{R}_0$ is \textbf{complete} iff $\forall$$x$ $\in$ $\mathcal{R}_0$, $\exists$$y$ = $\mathcal{F}$($x$) $|$ $y$ $\in$ $\mathcal{R}_f$ and  $\forall$$y$ $\in$ $\mathcal{R}_f$, $\exists$$x$ $\in$ $\mathcal{R}_0$ $|$ $y$ = $\mathcal{F}$($x$).
\end{definition}



%

\begin{definition}[Reachable set of a $\mathscr{NN}$-Layer] \label{def:reachlayer}
Let h: $\mathbb{R}^{j} \rightarrow{} \mathbb{R}^{p}$ be a $\mathscr{NN}-Layer$ as described in Definition \ref{def:nnLayer}. The reachable set $\mathcal{R}_{h}$, with input $\mathcal{X} \subset \mathbb{R}^{n}$ is defined as
    \begin{equation*}
    \centering
        \mathcal{R}_{h} = \{y~|~ y = h(x),  x \in \mathcal{X}\}.
    \end{equation*}
\end{definition}




Definition \ref{def:reachlayer}  applies to any discrete-time layer that is part of a GNODE, including, but not limited to the following supported $\mathscr{NN}-Layers$ supported in NNVODE: fully-connected layers with ReLU, tanh, sigmoid and leaky-ReLU activation functions, and convolutional-type layers such as batch normalization, 2-D convolutional, and max-pooling.
%

The reachability analysis of NODEs is akin to the general reachability problem for any continuous-time system modeled by an ODE. 
%
If we represent the NODE as a single layer $i$ of a GNODE with continuous dynamics described by ~\eqref{eq:omegalayer_func}, and assume that for a given initial state $z_0 \in \mathbb{R}^{j_i}$, the system admits a unique trajectory defined on $\mathbb{R}^+_0$, described by $\zeta(., z(t_0))$, then the reachable set of the given NODE can be characterized by Definition \ref{def:ODEreach}. 

\begin{definition}[Reachable set of a NODE] \label{def:ODEreach} Let $g$ be a \textbf{NODE} with solution $\zeta$(t; z$_0$) to ~\eqref{eq:omegalayer_func} for initial state z$_0$. The reachable set, $\mathcal{R}_{g}$ at $t = t_F$,  $\mathcal{R}_{g}(t_F)$, with initial set $\mathcal{R}_{0} \subset \mathbb{R}^{n}$ at time $t=t_0$ is defined as
    \begin{equation*}
    \centering
        \mathcal{R}_{g}(t_F) = \{ \zeta(t; z_0) \in \mathbb{R}^{n} ~|~ z_0 \in \mathcal{R}_{0}, t \in [0,t_f] \}.
    \end{equation*}
We also describe the reachable set for a time interval [$t_0$,$t_f$] as follows 
    \begin{equation*}
        \mathcal{R}_{g}([t_0,t_f]) := \bigcup_{t\in[t_0,t_f]}\mathcal{R}_{g}(t)
    \end{equation*}
\end{definition}

Now that we have outlined the general reachability problem of a NODE, whether we compute a single reachable set for the NODE at $t$ = $t_F$, or over an interval $t$ $\in$ [$t_0$,$t_F$] (computed by \emph{get$\_$time()}), depends on how the neural ODE was trained and the specific application of its use. However, the core computation remains the same. This is outlined in Algorithm \ref{alg:reach}.
%
%

\begin{algorithm}[h]
	\caption{Reachability analysis of a GNODE.}\label{alg:reach}
	\begin{algorithmic}[1]
        \small
		\Input{$\mathcal{F}, \mathcal{R}_0$}  ~{\scriptsize \textcolor{blue}{ $//$ GNODE, input set}}
		\Output{$\mathcal{R}_f$} ~{\scriptsize \textcolor{blue}{$//$ output reachable set}}
		\Procedure{\texttt{$\mathcal{R}_f$ = reach}}{$\mathcal{F}, \mathcal{R}_0$}
		\State N = $\mathcal{F}$.layers ~{\scriptsize \textcolor{blue}{ $//$ number of layers}}
		\For{\textit{i = 1 : N}} ~{\scriptsize \textcolor{blue}{ $//$ loop through every layer}}
		\State $\mathscr{L}_i$ = $\mathcal{F}$.layer(i)
		\If{$\mathscr{L}_i$ $is$ NODE }~{\scriptsize \textcolor{blue}{ $//$ check layer type}}
		\State $\textbf{t}_i$ = get$\_$time($\mathscr{L}_i$) ~{\scriptsize \textcolor{blue}{ $//$ get integration time bounds of layer $i$}}
		\State $\mathcal{R}_i$ =  reach$_{NODE}$($\mathscr{L}_i$,$\mathcal{R}_{i-1}$, $\textbf{t}_i$) ~{\scriptsize \textcolor{blue}{ $//$ reach set of layer $i$}, Definition \ref{def:ODEreach}}
		\Else \State $\mathcal{R}_i$ =  reach$_{\mathscr{NN}}$($\mathscr{L}_i$, $\mathcal{R}_{i-1}$) ~{\scriptsize \textcolor{blue}{ $//$ reach set of layer $i$}, Definition \ref{def:reachlayer}}
		\EndIf
        \EndFor
        \State $\mathcal{R}_f$ = $\mathcal{R}_N$ ~{\scriptsize \textcolor{blue}{ $//$ output reachable set}}
		\EndProcedure
              \end{algorithmic}
\end{algorithm}

\subsection{Reachability Methods}

In the previous sections, we defined the reachable set of a GNODE using a layer-by-layer approach. However, the computation of these reachable sets is defined by the reachability methods and set representations utilized in their construction. For instance, the same operations are not utilized to compute the reachable set for a fully-connected layer with a hyperbolic tangent activation function as with a fully convolutional layer. In the following section, we describe the set of methods in the NNVODE tool available to the community to compute the reachable set of each specific layer.

We begin with the NODE approaches, where we make a distinction based on the underlying dynamics. If the NODE is nonlinear, we make use of zonotope and polynomial-zonotope based methods that are implemented and available in CORA \cite{Althoff2015arch,althoff2013hscc}. If the NODE is purely linear, then we utilize the star set-based methods introduced in \cite{bak2017cav}, which are more scalable than other zonotope-based methods and possess soundness guarantees as well. 

For the $\mathscr{NN}$-Layers, there are several methods available, including zonotope-based and star-set based methods. However, we limited our implementation only using star sets to handle these layers, as this representation was demonstrated to be more computationally efficient and to enable tighter over-approximations of the derived reachable sets than zonotope methods \cite{tran2019fm}. Additionally, in using star set methods, we allow for the use of both approximate methods (\emph{approx-star}) and exact methods (\emph{exact-star}).
%
A summary of these methods and supported layers is depicted in Table \ref{tab:LayerSupport}, and for a complete description of the reachability methods utilized in our work, we refer the reader to \cite{tran2020cavtool} and the manual\footnote{NNV manual is available at: \url{https://github.com/verivital/nnv/blob/master/docs/manual.pdf}}\footnote{CORA manual (Release 2021) is available at: \url{https://tumcps.github.io/CORA/data/Cora2021Manual.pdf}}.

\begin{table}[h!]
    \caption{Layers supported in NNVODE and reachability sets and methods available.}
    \label{tab:LayerSupport}
    \centering
    \begin{adjustbox}{width=0.9\columnwidth, center}
    \begin{tabular}{c|l}
    \centering 
     & \textbf{Layer Type} -- \textbf{Set Rep. (method name)} \\
    \hline
    NODE & Linear -- Star-set ("direct") \cite{bak2017cav} \\ 
    NODE & Nonlinear -- Zonotope, Polynomial Zonotope * \cite{Althoff2015arch,althoff2013hscc} \\
    $\mathscr{NN}$-Layer & FC: linear, ReLU -- Star-set ("approx-star", "exact-star") \cite{tran2019fm} \\
    $\mathscr{NN}$-Layer & FC: leakyReLU, tanh, sigmoid, satlin -- Star-set ("approx-star") \cite{tran2020cavtool} \\ 
    $\mathscr{NN}$-Layer & Conv2D -- ImageStar ("approx-star", "exact-star") \cite{tran2020cavtool} \\ 
    $\mathscr{NN}$-Layer & BatchNorm -- ImageStar ("approx-star", "exact-star") \cite{tran2020cavtool} \\ 
    $\mathscr{NN}$-Layer & MaxPooling2D -- ImageStar ("approx-star", "exact-star") \cite{tran2020cavtool} \\ 
    $\mathscr{NN}$-Layer & AvgPooling2D -- ImageStar ("approx-star", "exact-star") \cite{tran2020cavtool} \\ 
    \hline 
    \end{tabular}
    \end{adjustbox}
    \begin{adjustwidth}{0.2cm}{}\scriptsize{* We support several methods available using Zonotope and PolyZonotopes, which includes user-defined fixed reachability parameters ("ZonoF", "PolyF") as well as adaptive reachability methods ("ZonoA" and "PolyA"), which require no prior knowledge on reach methods or systems to verify to produce relevant results.}
    \end{adjustwidth}
\end{table}

\noindent\textbf{Implementation.}
One of the key aspects of the verification of GNODEs is the proper encoding of the NODEs within reachability schemes. Depending on the software that is used, this process may vary and require distinct steps. As an example, some tools, like NNVODE, are simpler and allow for matrix multiplications within the definition of equations. However, for tools like Flow*, the set of steps required to properly encode this problem are more complex as it requires a definition for each individual equation of the state derivative. Thus, a more general conversion is needed, which is illustrated in the Appendix.

\section{Evaluation}\label{sec:eval}

Having described the details of our reachability definitions, algorithm, and implementation, we now present the experimental evaluation of our proposed work. 
We begin by presenting, a method and tool comparison analysis against
GoTube\footnote{GoTube can be found at \url{https://github.com/DatenVorsprung/GoTube}} \cite{gruenbacher2021gotube}, Flow*\footnote{Flowstar version 2.1.0 is available at \url{https://flowstar.org/}} \cite{Chen2013Flow*} and JuliaReach\footnote{JuliaReach can be found at \url{https://juliareach.github.io/}} \cite{bogomolov2019Juliareach}. 
Then, we present a case study of an Adaptive Cruise Control system, and conclude with an evaluation of the scalability of our techniques using a random set of architectures for dynamical system applications as well as a set of classification models for MNIST. The GNODE architectures for each benchmark can be found in the Appendix. To facilitate the reproducibility of our experiments, we set a timeout of 2 hours (7200 seconds) for each reach set computation\footnote{Code to reproduce all results can be found here: \url{https://github.com/verivital/nnv/tree/master/code/nnv/examples/Submission/FORMATS2022}}. All our experiments were conducted on a desktop with the following configuration: Intel Core i7-7700 CPU @ 3.6GHz 8 core Processor, 64 GB Memory, and 64-bit Ubuntu 16.04.3 LTS OS.

\subsection{Method and Tool Comparison}
We have implemented several methods within NNVODE, and the first evaluation consists of comparing the available methods for nonlinear NODEs, fixed-step zonotope and polynomial zonotopes (zono-F,poly-F) and adaptive zonotope and polynomial zonotope (zono-A,poly-A) based methods \cite{Althoff2015arch}. For all other $\mathscr{NN}$-Layers in the GNODEs, we use the star-set over-approximate methods. 
We considered multiple models, all inspired by the ILNODE representation that was introduced by Massaroli. They consist of a set of models with a varying number of augmented dimensions.
All these models are instances of the GNODE class presented in Definition \ref{def:neuralODE}, which present an architecture of the form $\mathscr{NN}$-Layers + NODE + $\mathscr{NN}$-Layers. In this context, we were concerned with how the methods scale with respect to the number of dimensions of each model.

\vspace{-1em}
\begin{table}[h!]
    \caption{Computation time of the reachability analysis of the Damped Oscillator benchmark. Results are shown in seconds with up to one decimal place. }
    \label{tab:doRes}
    \begin{adjustbox}{width=0.5\columnwidth,center}
    \begin{tabular}{cccccc}
    \toprule
    Aug. Dims & Zono-\emph{F} & Zono-\emph{A} & Poly-\emph{F} & Poly-\emph{A} \\ 
    \hline 
    0 & \textbf{34.0} & 574.5 & 201.7 & 654.0  \\ 
    1 & \textbf{146.4} & 4205.0 & 1573.9 & 3440.9 \\
    2 & \textbf{441.0} & -- & -- & -- \\
    \hline 
    \end{tabular}
    \end{adjustbox}%
\end{table}
\vspace{-4em}
\begin{figure}[!ht]
  \centering
    \subfigure[Aug. Dims = 0]{
        \includegraphics[trim=95 230 115 230, clip, width=0.23\columnwidth]{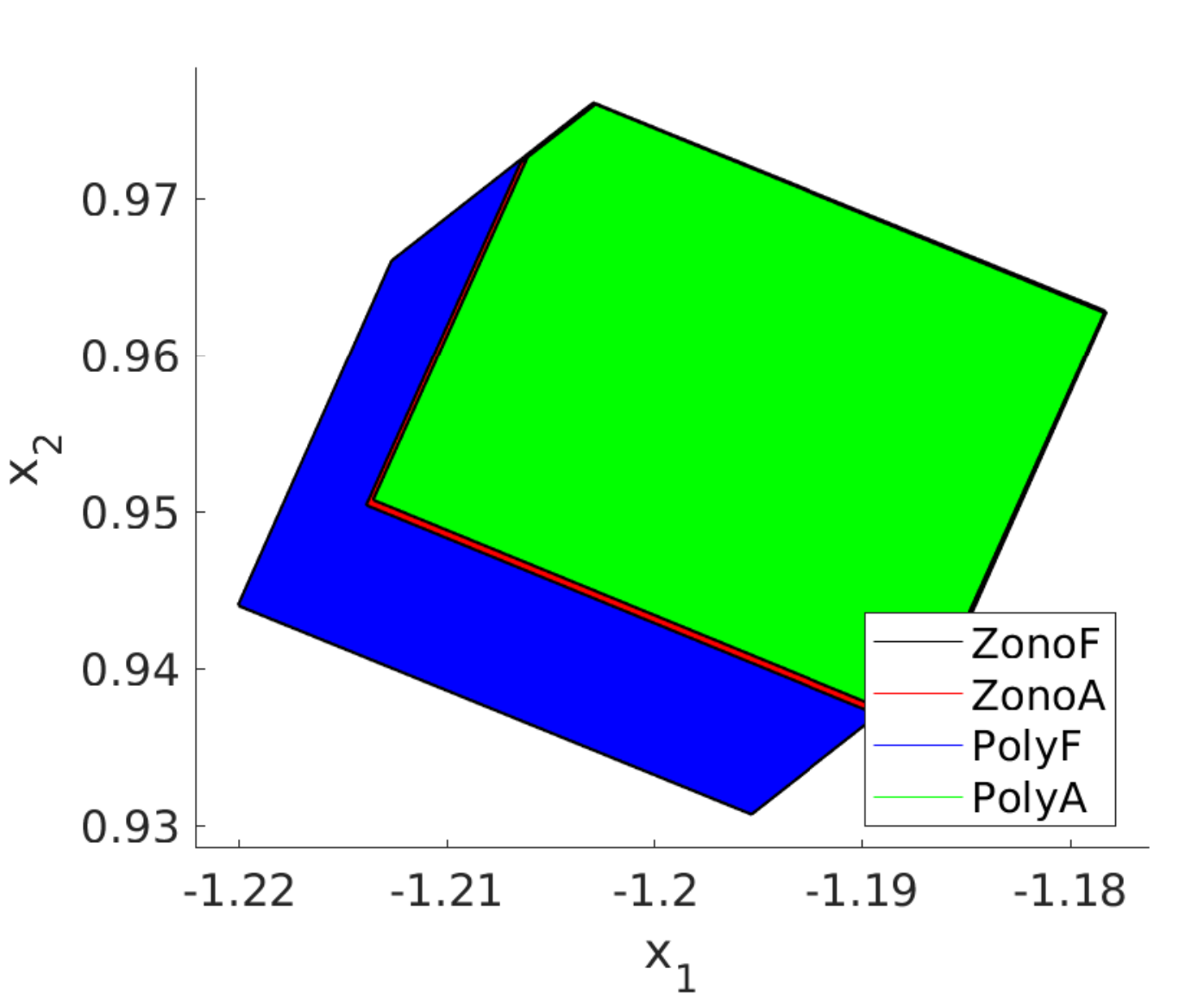}\label{fig:osc_0}}
    \subfigure[Aug. Dims = 0]{
        \includegraphics[trim=98 230 120 230, clip, width=0.23\columnwidth]{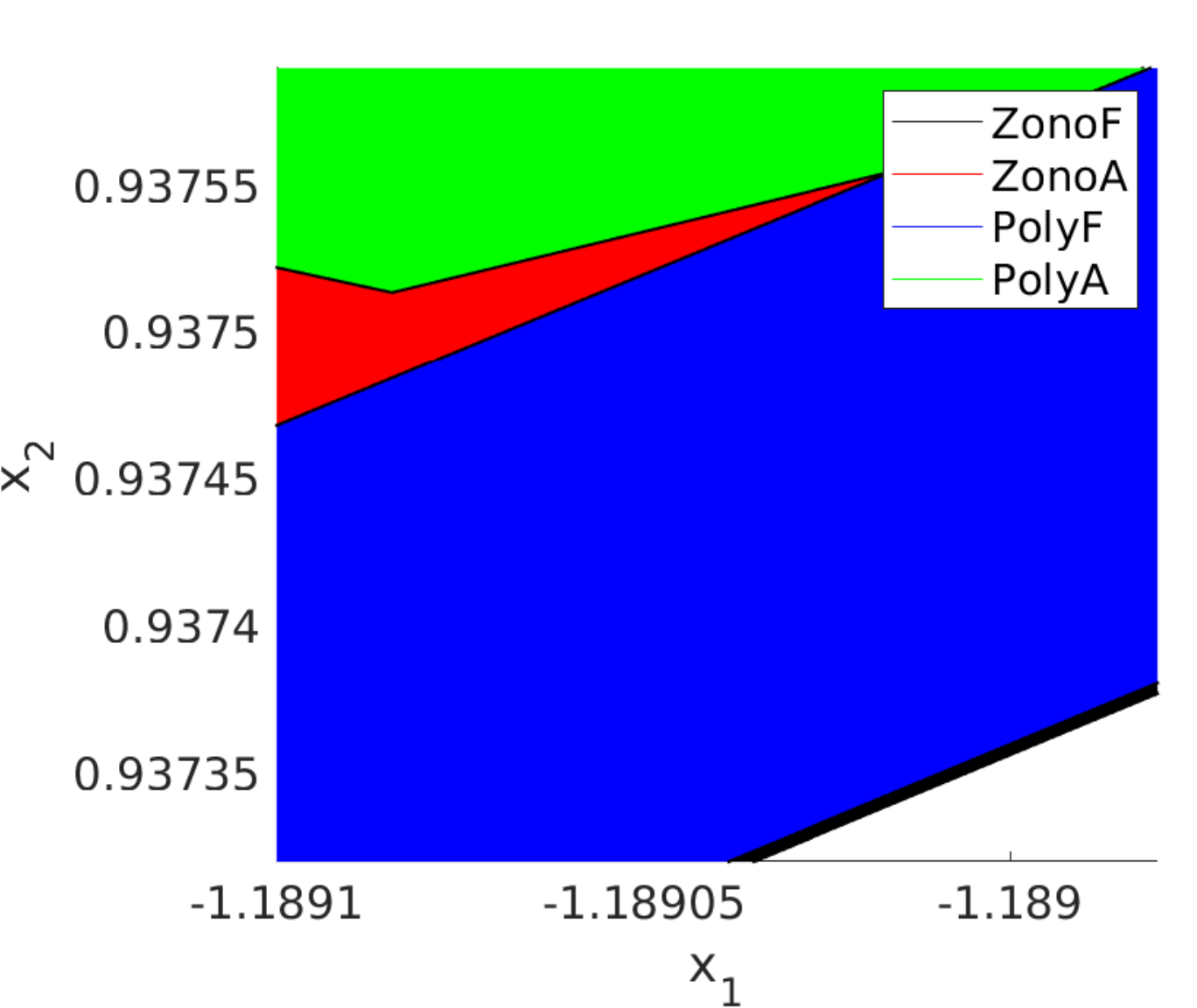}
        \label{fig:osc_0z}}
  \subfigure[Aug. Dims = 1]{
        \includegraphics[trim=98 230 115 240, clip, width=0.23\columnwidth]{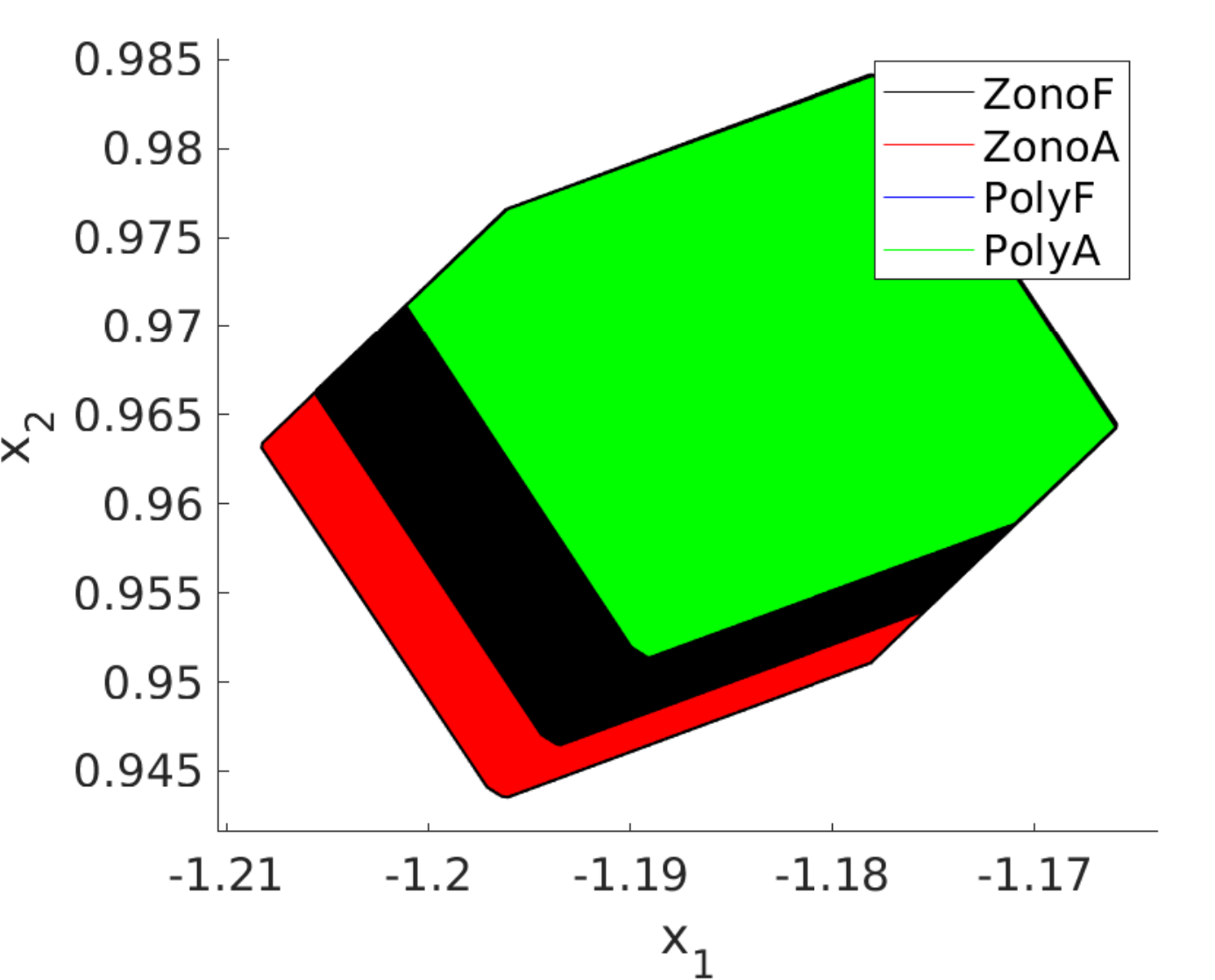}\label{fig:osc_1}}
    \subfigure[Aug. Dims = 1]{
        \includegraphics[trim=98 230 100 230, clip, width=0.23\columnwidth]{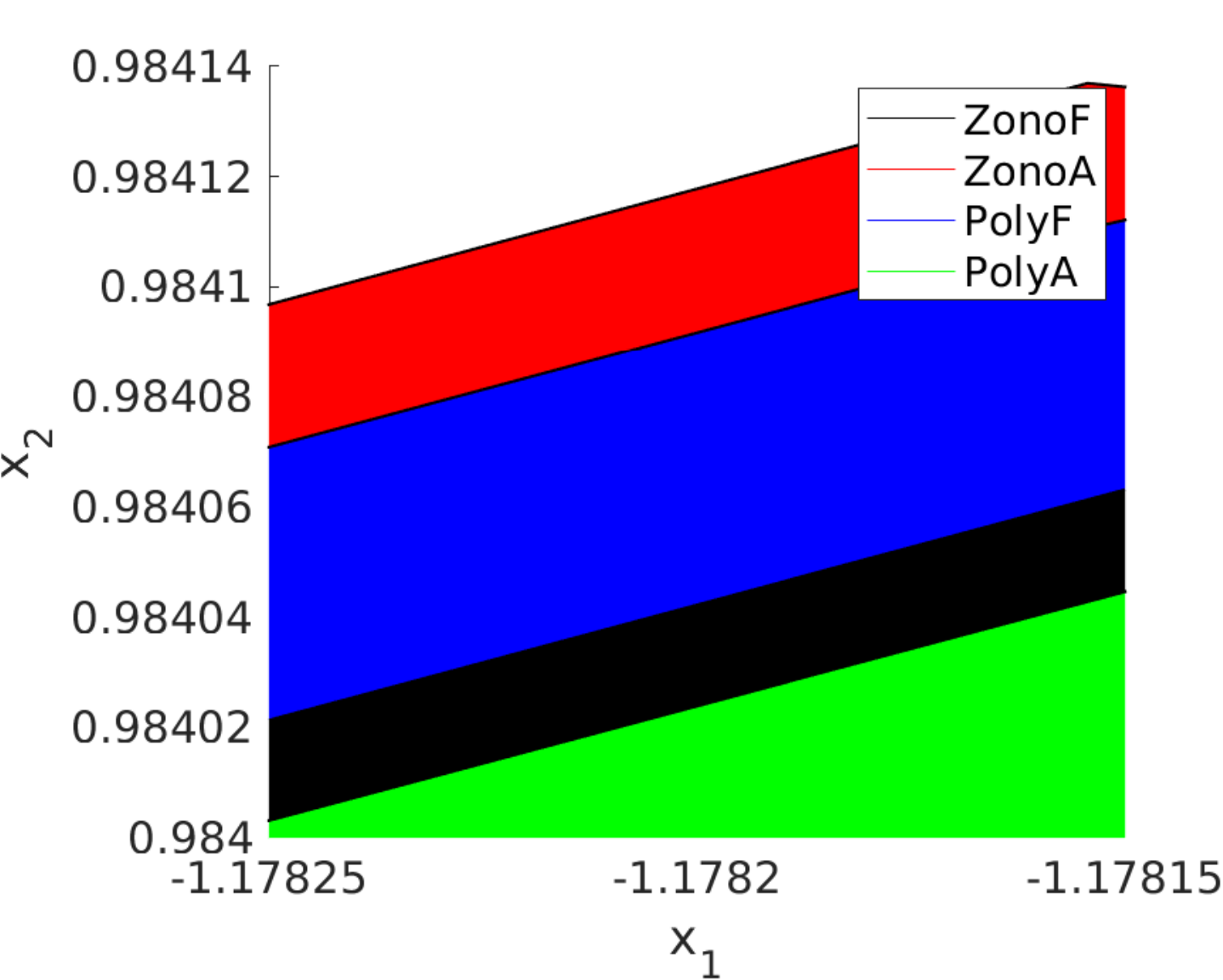}
        \label{fig:osc_1z}}
    \vspace{-1em}
    \caption{Reach sets comparison of the Damped Oscillator benchmark of the model with 0 and 1 augmented dimensions. Plots (a) and (c) show the reachable set at $t$ = 1s, and plots b) and d) show the zoomed-in reach sets to observe the minor size differences between the four methods: \textbf{\textcolor{red}{ZonoA}}, \textbf{ZonoF}, \textbf{\textcolor{blue}{PolyF}} and \textbf{\textcolor{Green}{PolyA}}.}
  \label{fig:damOsc_compare}
\end{figure}
\vspace{-1em}

In Table \ref{tab:doRes}, we observe that the Zono-$F$ method is the fastest across all models, while the adaptive methods are the slowest. Moreover, only Zono-$F$ is able to complete the reach set computation for all three models, while the other methods time out. In terms of the size of the computed reach sets, we compare the last reach set obtained in Figure \ref{fig:damOsc_compare}. In the subplots \ref{fig:osc_0z} and \ref{fig:osc_1z} we see the zoomed in reach sets, and observe that the Poly-$A$ method computes the smallest over-approximate reach set across both experiments.
Based on these results, in all subsequent experiments, we use the \emph{zono-F} method for nonlinear NODEs, the \emph{direct} method for linear NODEs, and the over-approximate star-set methods for all the $\mathscr{NN}$-Layers.

\begin{figure}[!ht]
  \centering
    \subfigure[Spiral$_{L_3}$]{
        \includegraphics[trim=98 240 120 250, clip, width=0.23\columnwidth]{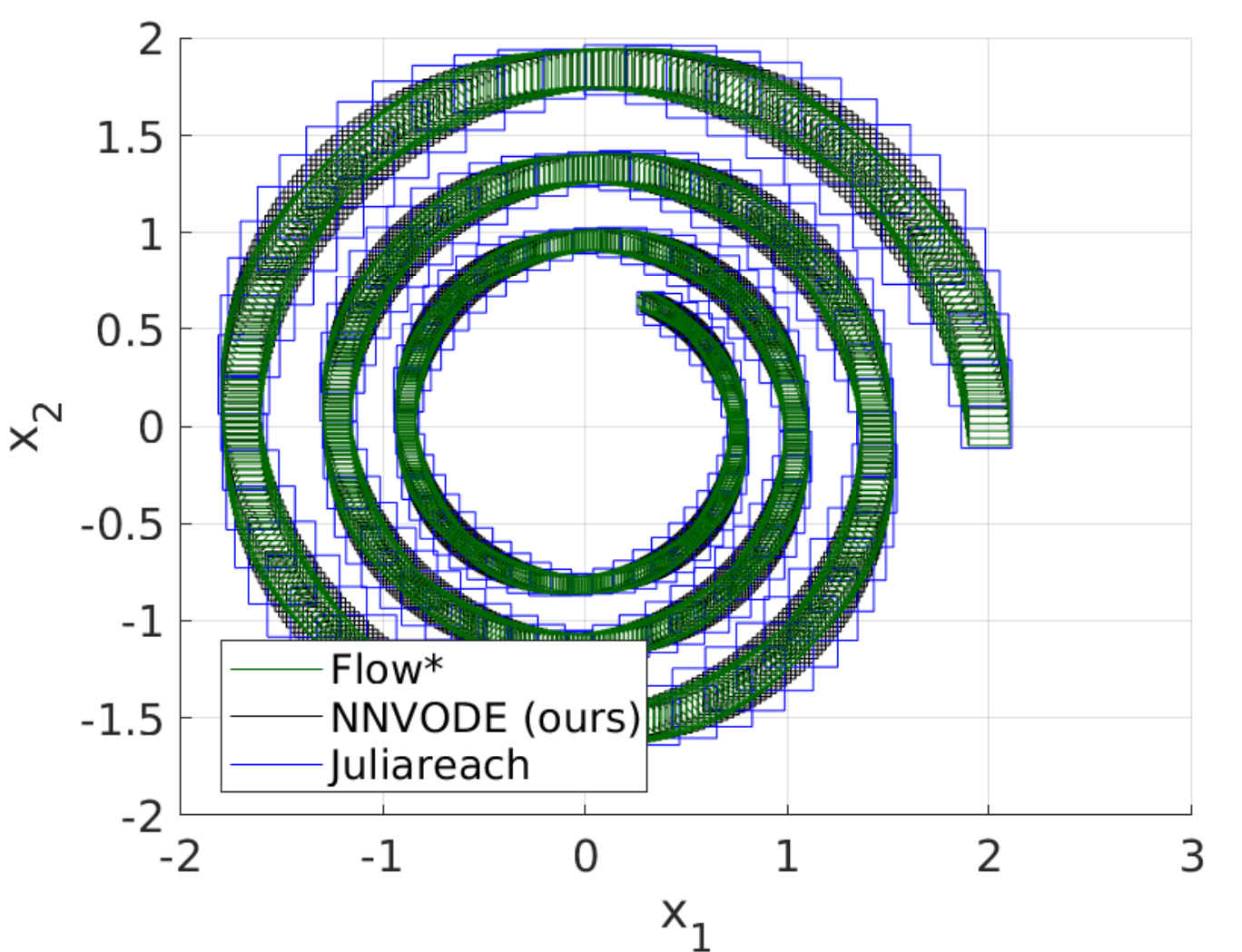}\label{fig:spiral_la}}
    \subfigure[Spiral$_{NL_3}$]{
        \includegraphics[trim=100 230 120 240, clip, width=0.23\columnwidth]{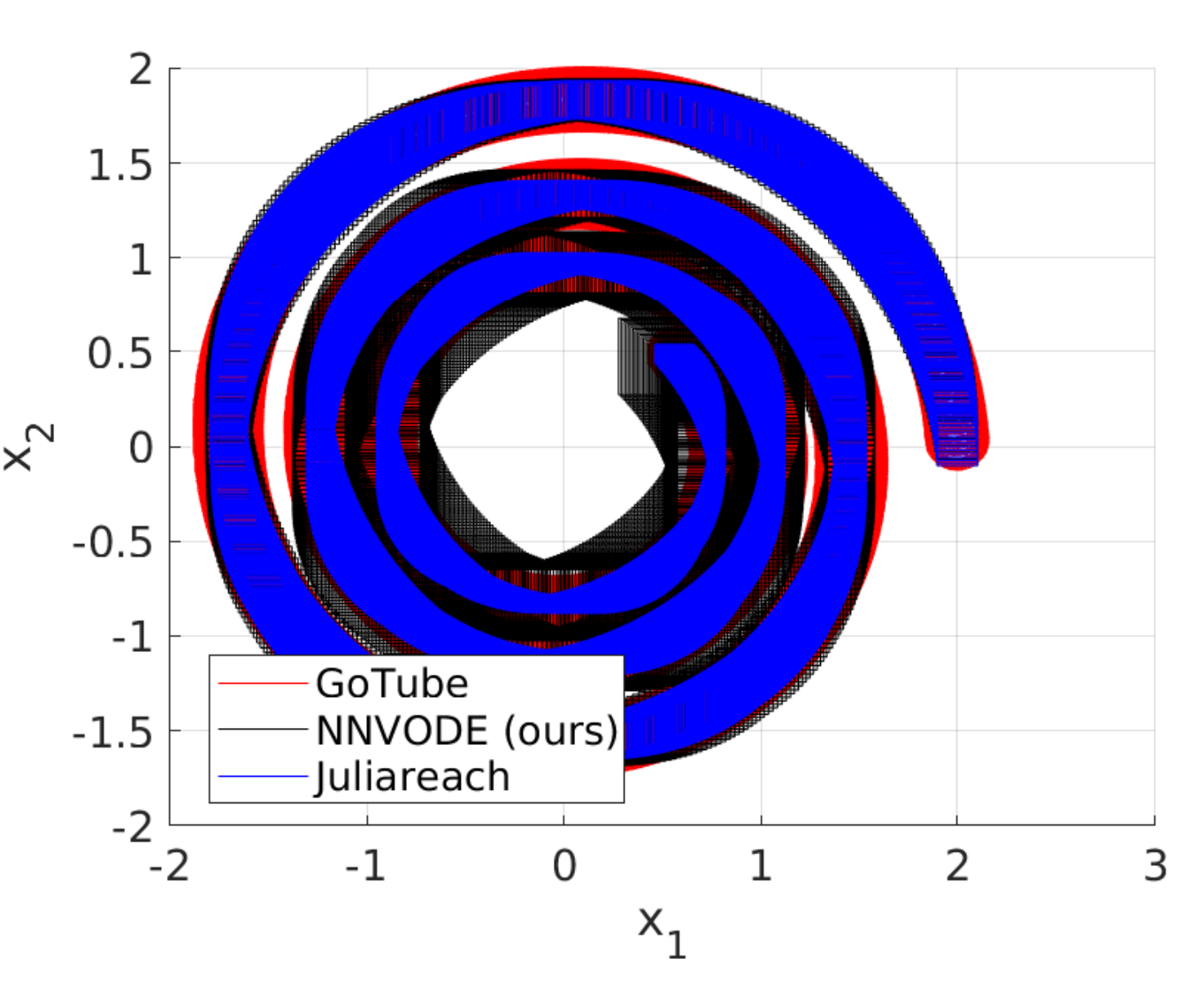}
        \label{fig:spiral_nla}} 
     \subfigure[FPA$_3$]{
        \includegraphics[trim=100 240 130 250, clip, width=0.23\columnwidth]{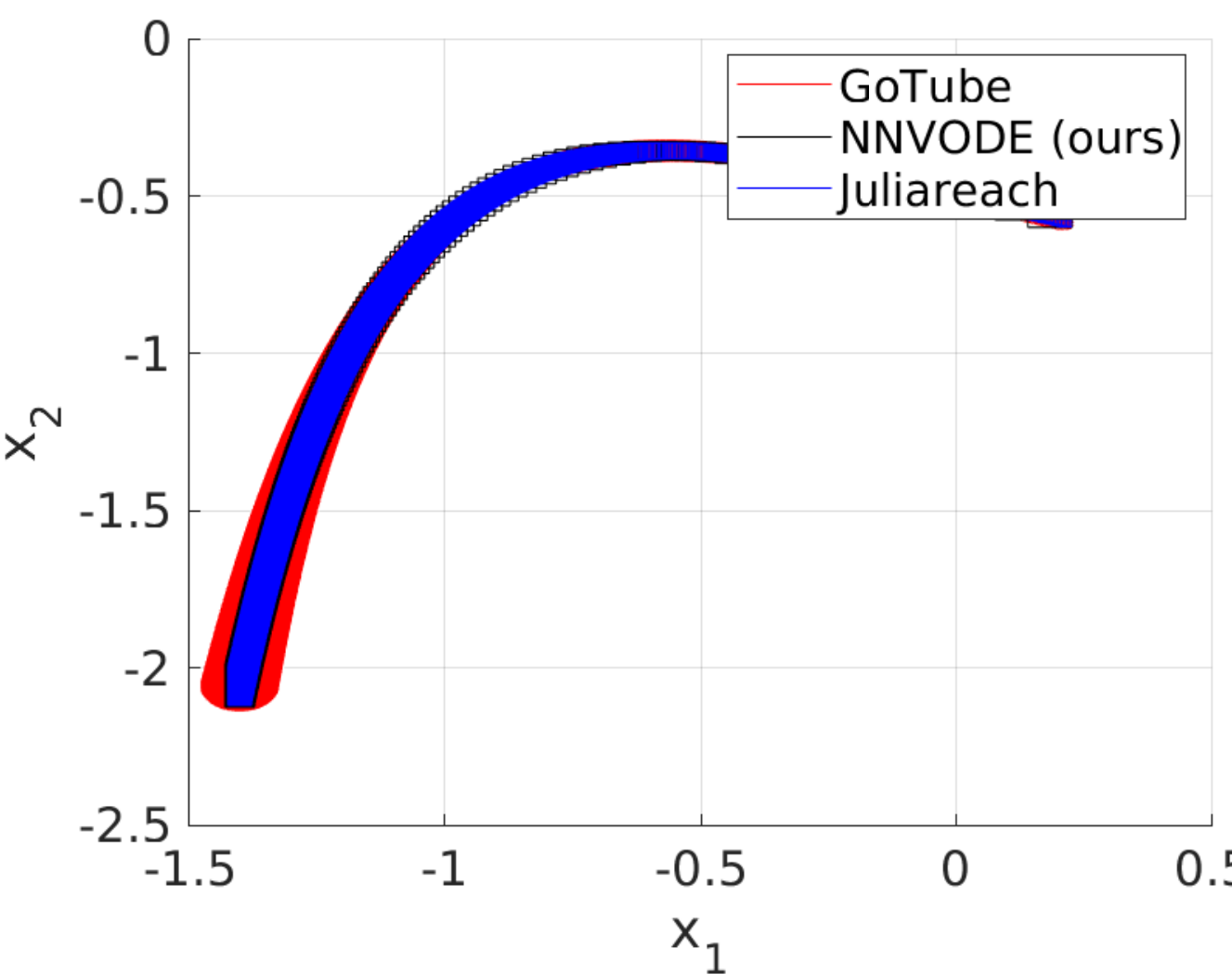}\label{fig:fpa_a}}
    \subfigure[Cartpole$_3$]{
        \includegraphics[trim=95 230 115 240, clip, width=0.23\columnwidth]{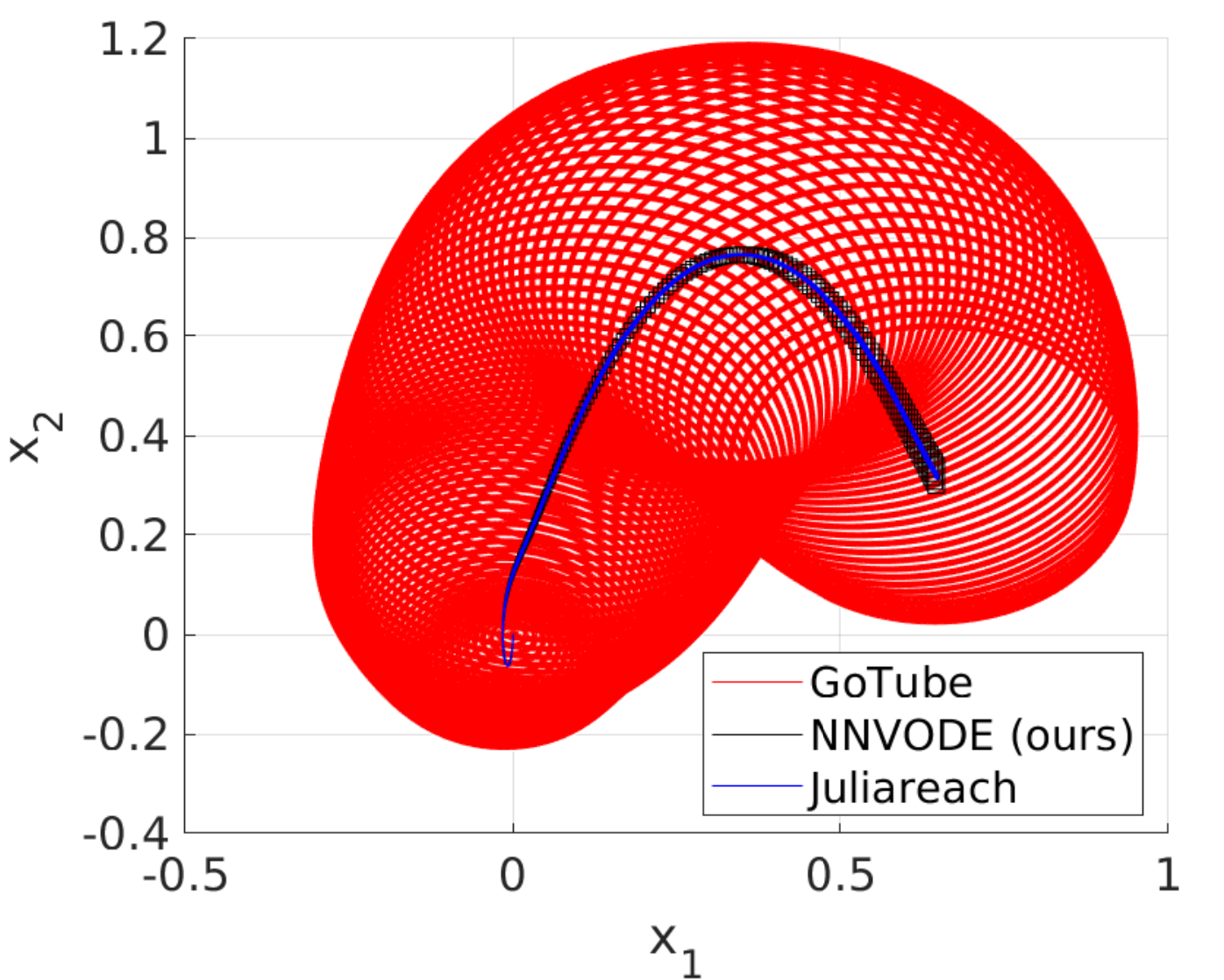}\label{fig:cartpole_a}}
    \subfigure[Spiral$_{L_3}$]{
        \includegraphics[trim=98 230 120 240, clip, width=0.23\columnwidth]{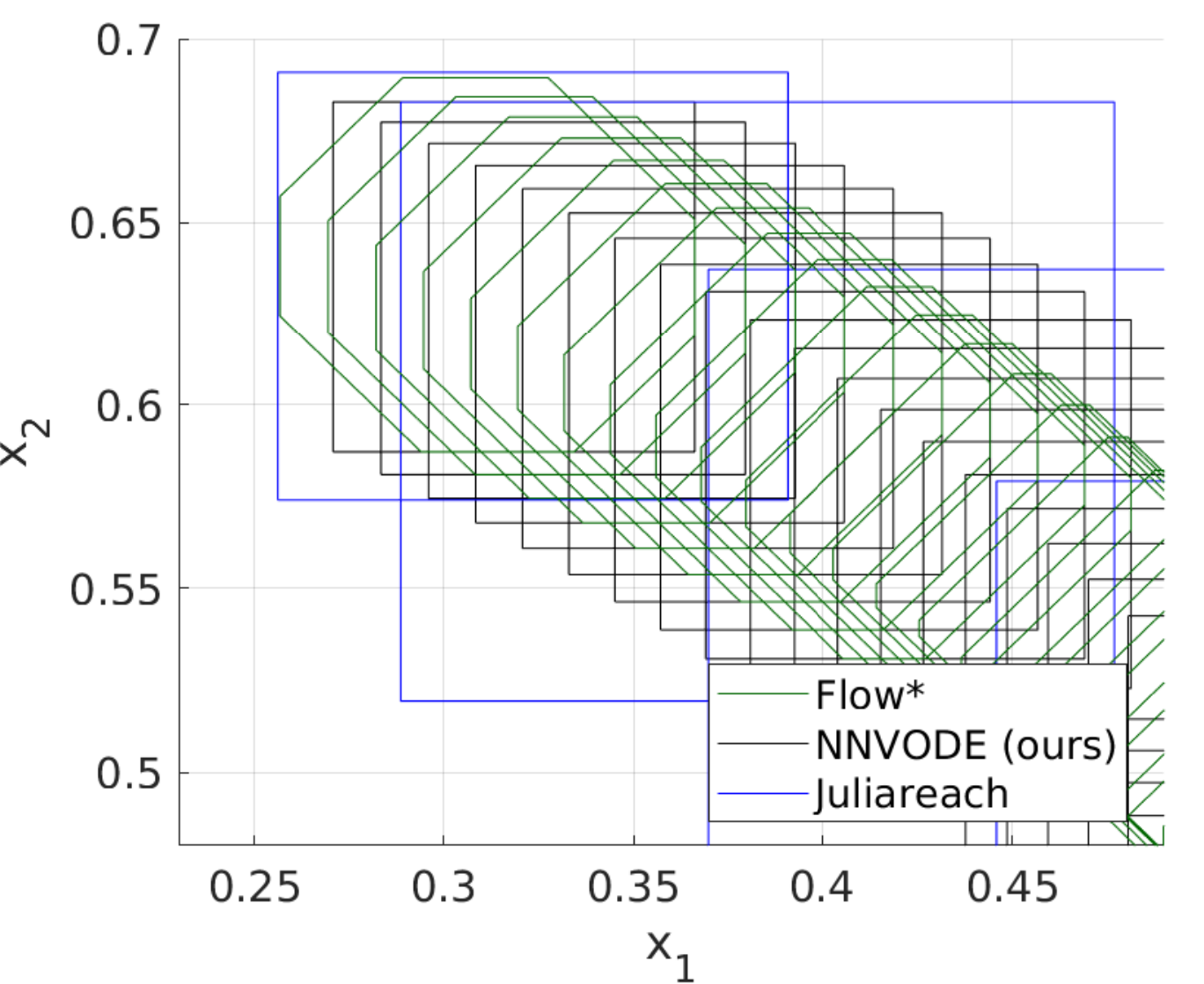}\label{fig:spiral_lc}}
    \subfigure[Spiral$_{NL_3}$]{
        \includegraphics[trim=100 230 120 240, clip, width=0.22\columnwidth]{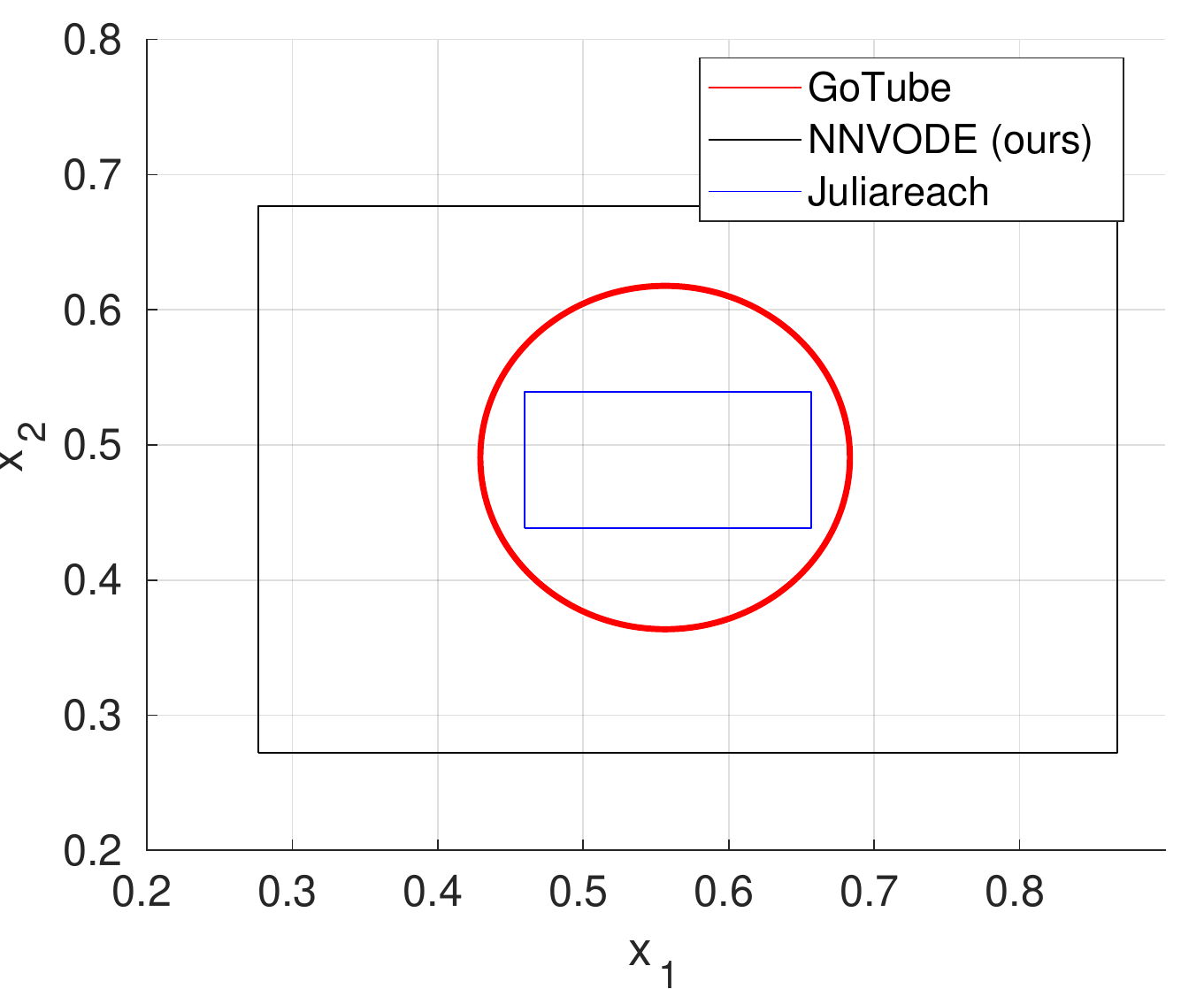}
        \label{fig:spiral_nlb}} 
    \subfigure[FPA$_3$]{
        \includegraphics[trim=95 230 120 240, clip, width=0.23\columnwidth]{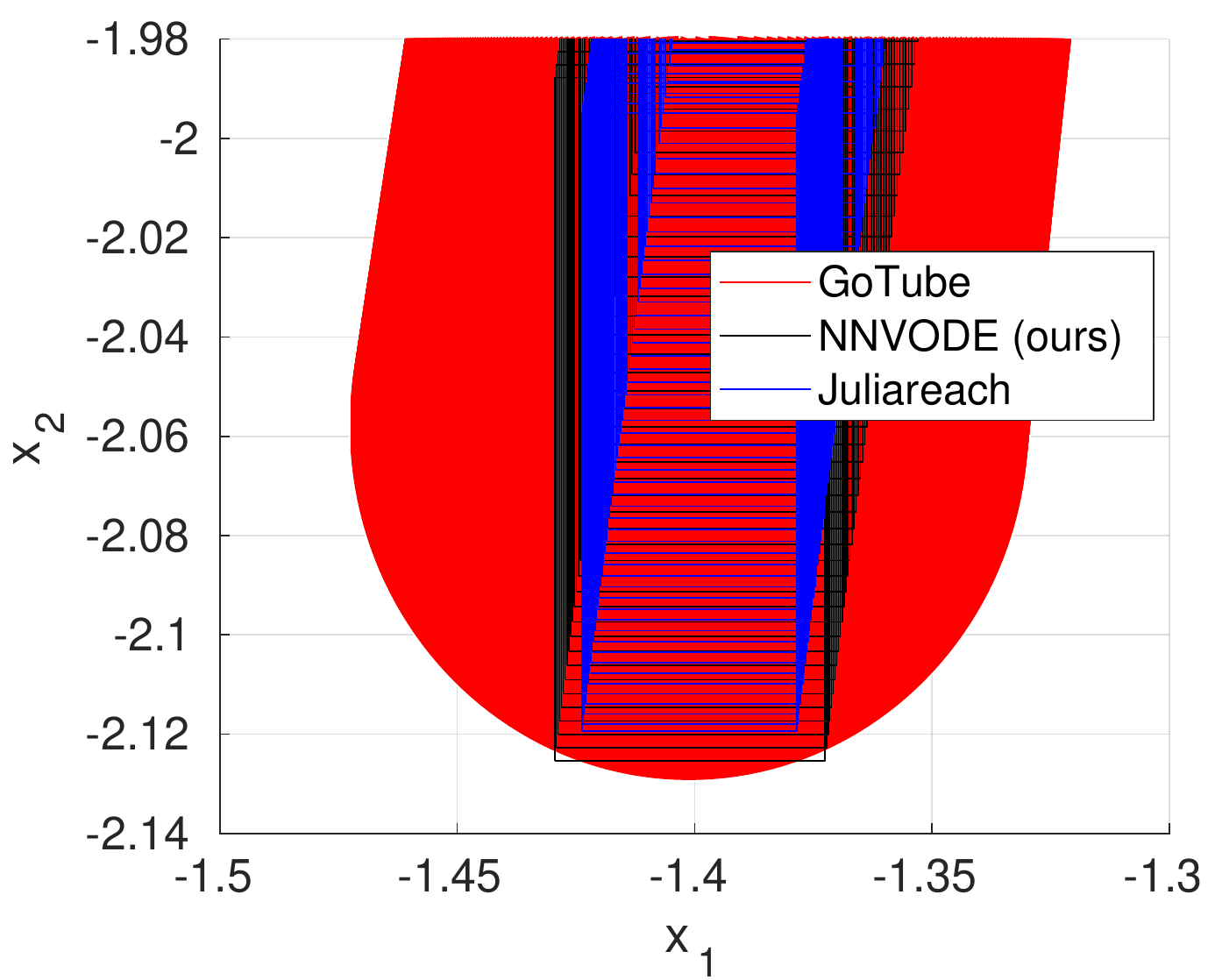}
        \label{fig:fpa_b}} 
    \subfigure[Cartpole$_3$]{
        \includegraphics[trim=95 230 115 240, clip, width=0.23\columnwidth]{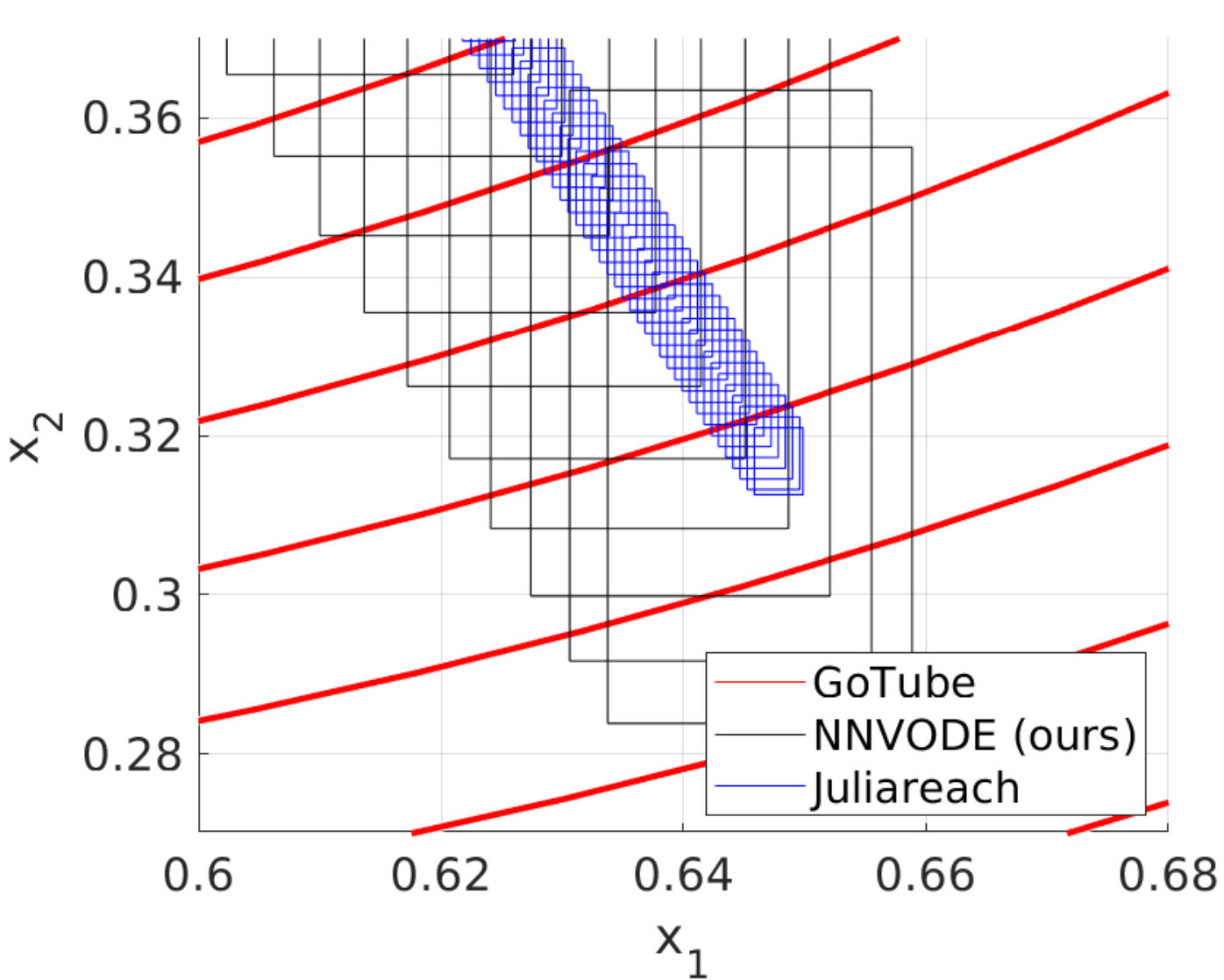}
        \label{fig:cartpole_b}}
    \vspace{-1em}
  \caption{NODE reach set comparisons. The top 4 figures ([$a$,$d$]) show the complete reachable sets of each benchmark, while the bottom 4 correspond to the zoomed-in reach sets ($e$, $g$) and to the zoomed-in figure of the last reach set ($f$,$h$). The figures show the computed reach sets of \textbf{\textcolor{red}{GoTube}}, \textbf{NNVODE}, \textbf{\textcolor{blue}{JuliaReach}} and \textbf{\textcolor{OliveGreen}{Flow*}}.}
  \label{fig:res1compare}
\end{figure}

The next part of our evaluation consists of  comparing NNVODE's methods for NODEs to those of Flow* \cite{Chen2013Flow*}, Gotube \cite{gruenbacher2021gotube} and JuliaReach \cite{bogomolov2019Juliareach} across a collection of benchmarks that include a linear and nonlinear 2-dimensional spiral \cite{chen2018neural}, a Fixed-Point Attractor (FPA) \cite{musau2018arch} and a controlled cartpole \cite{gruenbacher2020lagrangian}. The computationn reachability results are displayed in Table \ref{tab:res1} with the intention to characterize major differences between tools, i.e., to show some tools are 10$\times$ to 20$\times$ faster than others for some benchmarks. It is worth noting, that we are not experts on every tool that we considered. Thus it may be possible to optimize the reachable set computation for each benchmark with depending on the tool. However, we did not do so. Instead, we attempted around 3 to 5 different parameter combinations for each benchmark, and used the best results we could obtain when comparing against the other tools. The details can be found in the Appendix.
The first three rows correspond to the linear spiral 2D model, and the subsequent three rows to the nonlinear one. The first set of observations that can be made in this context is that Flow* times out on all problems involving nonlinear neural ODEs, and that GoTube cannot obtain a solution to the reachability problem for the linear model. In terms of computation time, the results vary. Flow* is the fastest for the linear Spiral 2D model, regardless of the size of the initial set. In general, JuliaReach and NNVODE are much faster than GoTube, with JuliaReach being the fastest tool across the board. Additionally, there is not a significant difference between NNVODE and JuliaReach in the treatment of the nonlinear spiral model and FPA models. However, JuliaReach is an order of magnitude faster than NNVODE and two orders of magnitude faster than GoTube on the cartpole benchmark.
In Figure \ref{fig:res1compare} we display a subset of the reachability results from Table \ref{tab:res1} where we observe that
JuliaReach is able to compute smaller over-approximations of the reachable set on all the benchmarks except for the linear spiral model. Notably, in most cases, 
GoTube computes the largest over-approximation, and this effect grows far more significantly than the other tools when the complexity of the model increases. This can be  observed in Figures \ref{fig:cartpole_a} and \ref{fig:cartpole_b}.

\begin{table}[h!]
    \caption{Results of the reachability analysis of the NODE benchmarks. All results are shown in seconds. TH stands for Time Horizon and $\delta_{\mu}$ to the input uncertainty.}
    \label{tab:res1}
    \centering
    \addtolength{\tabcolsep}{5pt}
    \begin{adjustbox}{width=0.75\columnwidth,center}
    \begin{tabular}{c|cc|cccc}
    \centering 
    Name & TH ($s$) & $\delta_{\mu}$ & Flow* & GoTube & JuliaReach & NNVODE (ours) \\ 
    \hline 
    Spiral$_{L_1}$ & 10 & 0.01 & \textbf{4.0} & -- & 10.2 & 8.0 \\ 
    Spiral$_{L_2}$ & 10 & 0.05 & \textbf{3.7} & -- & 7.2 & 7.4 \\
    Spiral$_{L_3}$ & 10 & 0.1 & \textbf{3.7} & -- & 7.2 & 7.3 \\ \hline
    Spiral$_{NL_1}$ & 10 & 0.01 & -- & 106.4 & 58.9 & \textbf{47.4} \\ 
    Spiral$_{NL_2}$ & 10 & 0.05 & -- & 106.7 & \textbf{41.7} & 46.2 \\ 
    Spiral$_{NL_3}$ & 10 & 0.1 & -- & 106.5 & \textbf{41.9} & 46.2 \\ \hline
    FPA$_1$ & 0.5 & 0.01 & - & 13.6 & 1.9 & \textbf{1.3}  \\ 
    FPA$_2$ & 2.5 & 0.01 & - & 42.4 & \textbf{5.6} & 6.6 \\ 
    FPA$_3$ & 10.0 & 0.01 & - & 140.2 & 28.4 & \textbf{7.5} \\ \hline
    Cartpole$_1$ & 0.1 & 1$e$-4 & - & 183.0 & \textbf{3.2} & 67.9  \\ 
    Cartpole$_2$ & 1.0 & 1$e$-4 & - & 1590.9 & \textbf{13.8} & 404.3  \\
    Cartpole$_3$ & 2.0 & 1$e$-4 & - & 3065.7 & \textbf{35.5} & 834.7  \\
    \hline 
    \end{tabular}
    \end{adjustbox}
\end{table}


\subsection{Case Study: Adaptive Cruise Control (ACC)}

This case study was selected to evaluate an original NNCS benchmark used in all the AINNCS ARCH-Competitions \cite{ARCH2021ainncs,ARCH20AINNCS,manzanas2019arch_ainncs,tran2020cavtool} against NNCS with NODEs as dynamical plants learned from simulation data using 3$^{rd}$ order NODEs \cite{norcliffe2020sonode}. We demonstrate the verification of the ACC with different GNODEs learned as the plant model of the ACC and compare against the original benchmark, while using the same NN controller across all three models.  
The details of the original ACC NNCS benchmark can be found in \cite{tran2020cavtool}, and the architectures of the third order neural ODEs can be found in the Appendix. 

%
\begin{figure}[!ht]
  \centering
    \subfigure[Original]{
        \includegraphics[trim = 0 0 0 0,clip, width=0.31\columnwidth]{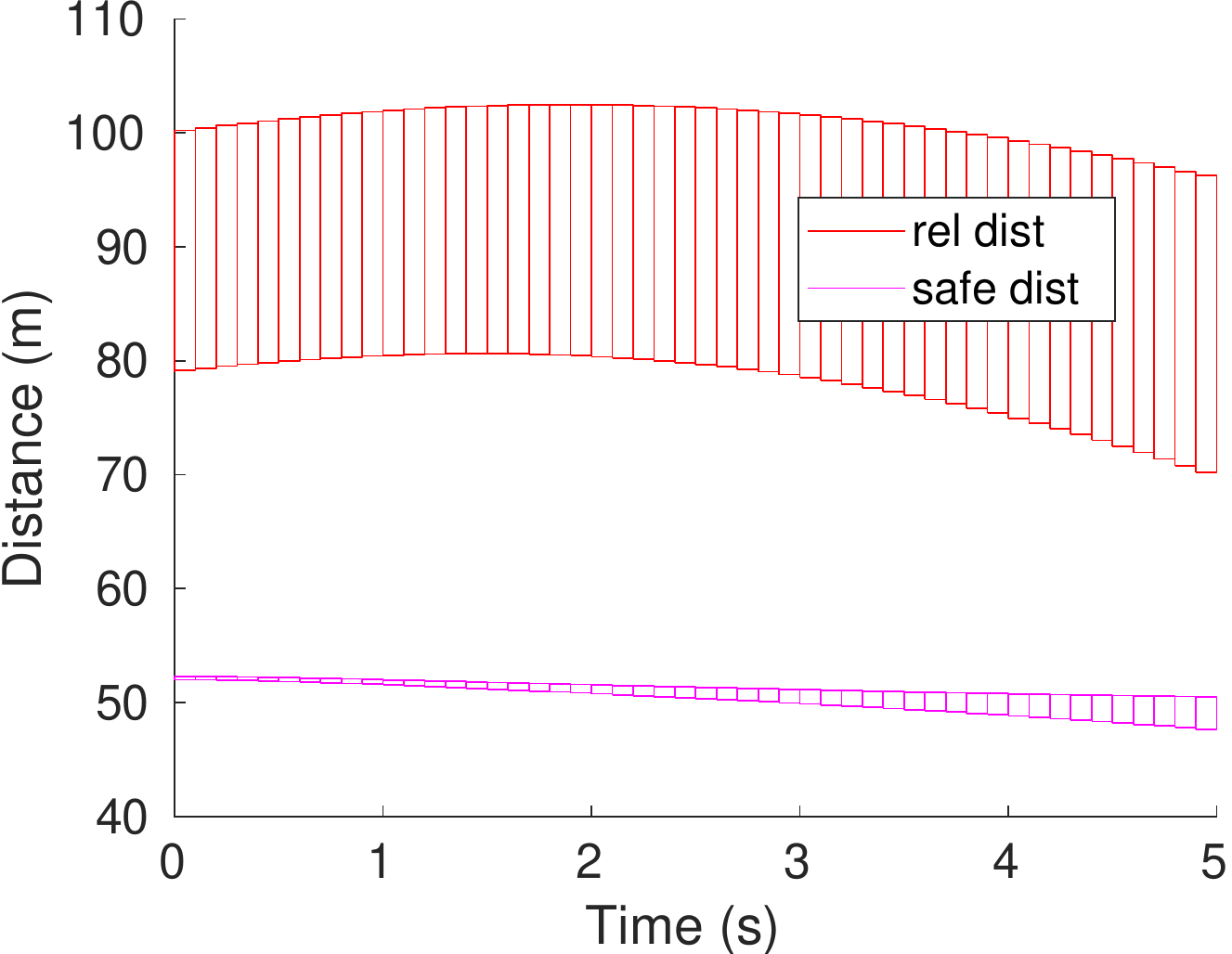}\label{fig:acc_orig}}
    \subfigure[Linear NODE]{
        \includegraphics[trim = 0 0 0 0,clip, width=0.31\columnwidth]{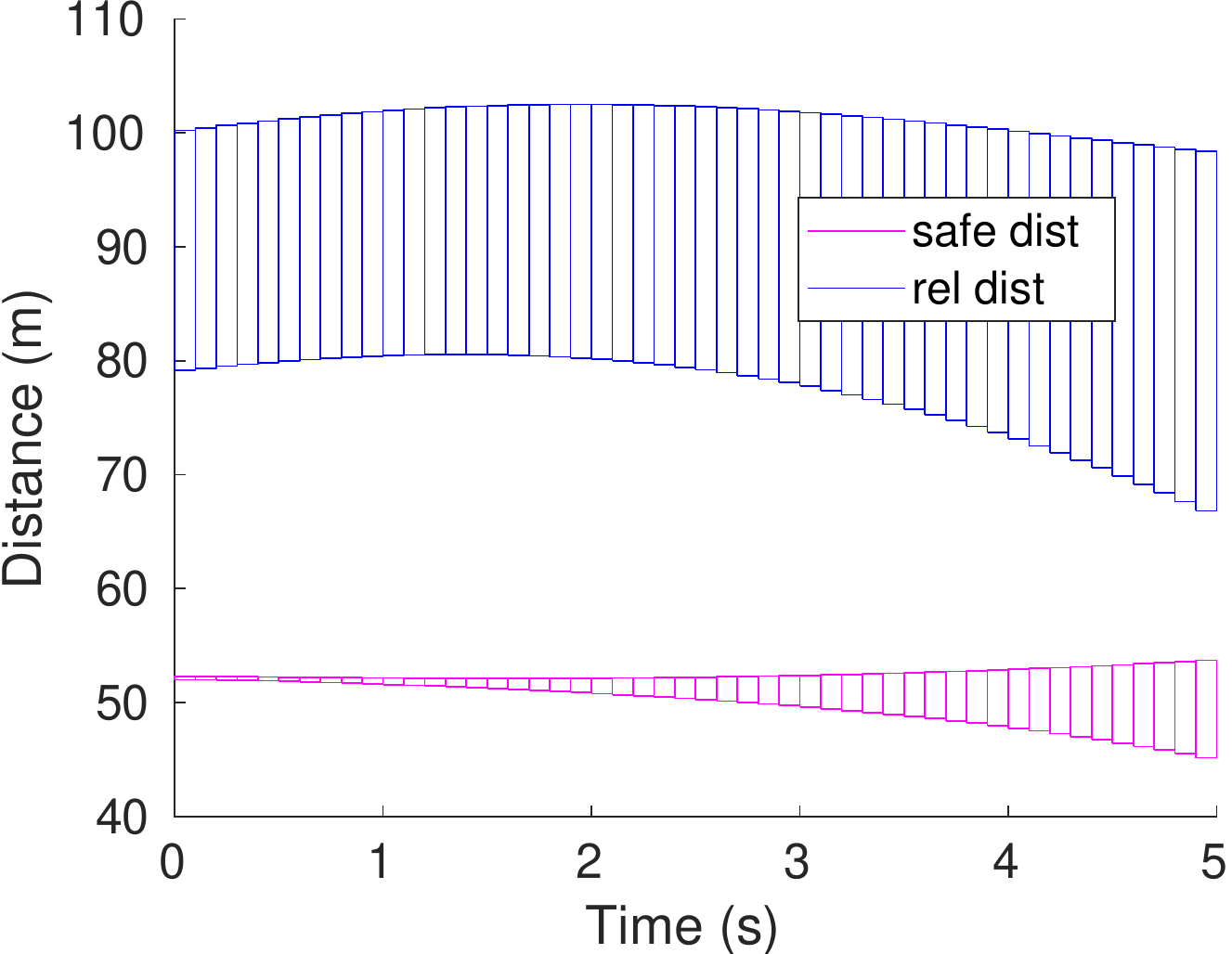}\label{fig:acc_lin}}
    \subfigure[Nonlinear NODE]{
        \includegraphics[trim = 0 0 0 0,clip, width=0.31\columnwidth]{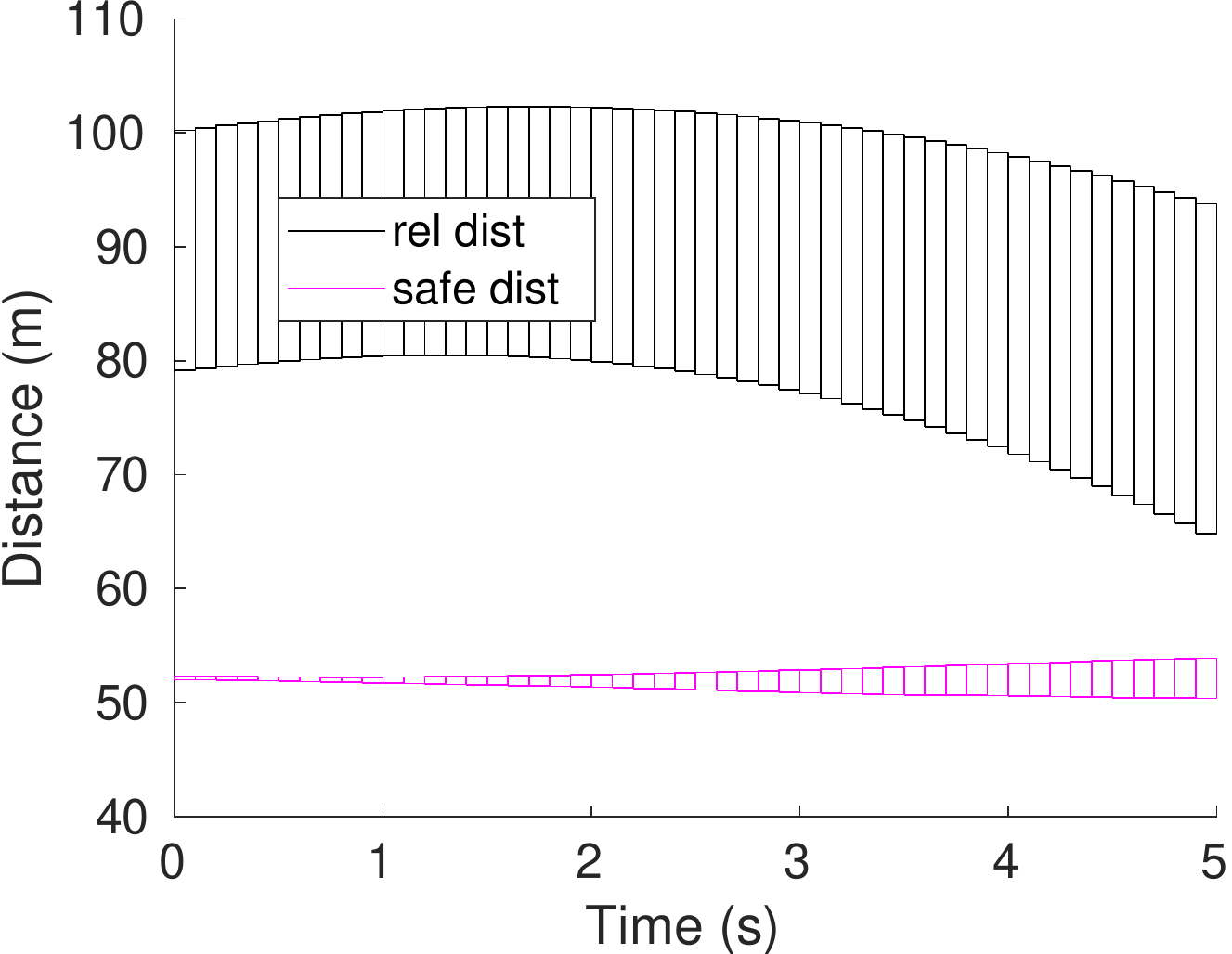}\label{fig:acc_nl}}
    \vspace{-1em}
  \caption{Adaptive Cruise Control comparison. In \textcolor{red}{red} we display the relative distance of the original plant, in \textcolor{blue}{blue} the linear NODE, in \textbf{black} the nonlinear NODE, and in \textcolor{magenta}{magenta}, the safe distance.}
  \label{fig:acc}
\end{figure}

We considered all three models using the same initial conditions and present the results in Figure \ref{fig:acc}. The reachable sets obtained from all three plants are largely similar, and we can guarantee that all the models are safe since the intersection between the safe distance and the relative distance is empty. When we consider the size of the reachable sets, one can see that the original model returns the smallest reach set, whereas the linear model boasts the largest one. In all, the biggest difference between the plant dynamics is the computation time. Here, the linear 3$^{rd}$ order NODE boasts the fastest computation time of 0.86 seconds, whereas the original plant takes 14.9 seconds, and the nonlinear 3$^{rd}$ order NODE takes 998.7 seconds. 

\subsection{Classification NODEs}

Our second set of experiments considered performing a robustness analysis for a set of MNIST classification models with only fully-connected $\mathscr{NN}$-Layers and other 3 models with convolutional layers as well. There is one linear NODE in each model, and we vary the number of parameters and states across all of them to study the scalability of our methods. We evaluate the robustness of these models under an L$_{\infty}$ adversarial perturbation of $\epsilon$ = $\{$0.05, 1, 2$\}$ over all the pixels, and $\epsilon$ = $\{$2.55, 12.75, 25.5$\}$ over a subset of pixels (80). \footnote{Adversarial perturbations are applied before normalization, pixel values z$_p$ $\in$ [0,255]. } The complete evaluation of this benchmark consists of a robustness analysis using 50 randomly sampled images for each attack. We compare the number of images the neural ODEs are robust to, as well as the total computation time to run the reachability analysis for each model in Table \ref{tab:mnistRes}.

\begin{definition}[Robustness]\label{def:robustness}
    Given a classification-based GNODE $\mathcal{F}(z)$, input $z \in \mathbb{R}^j$, perturbation parameter $\epsilon \in \mathbb{R}$ and an input set $Z_p$ containing $z_p$ such that $Z_p = \{z: ||z - z_p|| \leq \epsilon\}$ that represents the set of all possible perturbations of $z$. The neural ODE is locally \textbf{robust} at $z$ if it classifies all the perturbed inputs $z_p$ to the same label as $z$, i.e., the system is \textbf{robust} if $\mathcal{F}(z_p) = \mathcal{F}(z)$ for all $z_p \in Z_p$.
\end{definition}
\begin{table}[h!]
    \caption{Robustness analysis of MNIST classification GNODEs under L$_{\infty}$ adversarial perturbations. The accuracy and robustness results are described as percentage values between 0 and 1, and the time computation corresponds to the average time to compute the reachable set per image. Columns 3-8 corresponds to $\epsilon$ = \{0.5,1,2\} over all pixels in the image, columns 9-14 corresponds to $\epsilon$ = \{2.55,12.75,25.5\} attack over a subset of pixels (80) n each image.}
    \label{tab:mnistRes}
    \centering
    \begin{adjustbox}{width=\columnwidth,center}
    \addtolength{\tabcolsep}{3pt} 
    \begin{tabular}{lc|cc|cc|cc|cc|cc|cc|} 
    \centering 
     & & \multicolumn{6}{c|}{$\ell$$_{\infty}$} & \multicolumn{6}{c|}{$\ell$$_{\infty}$ $_{(80)}$} \\ 
    \cline{3-14}
     & & \multicolumn{2}{c|}{\textbf{0.5}} & \multicolumn{2}{c|}{\textbf{1}} & \multicolumn{2}{c|}{\textbf{2}} & \multicolumn{2}{c|}{\textbf{2.55}} & \multicolumn{2}{c|}{\textbf{12.75}} & \multicolumn{2}{c|}{\textbf{25.5}}\\ 
    \cline{3-14}
     \textbf{Name} & \textbf{Acc.} & \textbf{Rob.} & \textbf{T(s)} & \textbf{Rob.} & \textbf{T(s)} & \textbf{Rob.} & \textbf{T(s)} & \textbf{Rob.} & \textbf{T(s)} & \textbf{Rob.} & \textbf{T(s)} & \textbf{Rob.} & \textbf{T(s)} \\
    \hline
    FNODE$_{S}$ & 0.9695 & 1 & 0.0836 & 0.98 & 0.1062 & 0.98 & 0.1186 & 1 & 0.0192 & 0.98 & 0.0193 & 0.98 & 0.0295\\ 
    FNODE$_{M}$ & 0.9772 & 1 & 0.0948 & 1 & 0.1177 & 0.98 & 0.2111 & 1 & 0.0189 & 1 & 0.0225 & 0.98 & 0.0396 \\ 
    FNODE$_{L}$ & 0.9757 & 1 & 0.1078 & 1 & 0.1674 & 0.96 & 0.5049 & 1 & 0.0187 & 1 & 0.0314 & 0.96 & 0.0709 \\ 
    CNODE$_{S}$ & 0.9706 & 0.98 & 13.95 & 0.96 & 15.82 & 0.86 & 17.58 & 0.98 & 1.943 & 0.94 & 3.121 & 0.78 & 4.459 \\ 
    CNODE$_{M}$ & 0.9811 & 1 & 176.5 & 1 & 193.8 & 1 & 293.8 & 1 & 21.45 & 1 & 83.28 & 1 & 182.0 \\ 
    CNODE$_{L}$ & 0.9602 & 1 & 1064 & 1 & 1089 & 1 & 1736 & 1 & 234.6 & 1 & 522.7 & 1 & 779.3 \\ 
    \hline 
    \end{tabular}
    \addtolength{\tabcolsep}{-3pt} 
    \end{adjustbox}
\end{table}
\vspace{-1em}

Finally, we performed a scalability study using a set of random GNODE architectures with multiple NODEs. Here, we focus on the nonlinear methods for both the $\mathscr{NN}$-Layers and the NODEs and evaluate how our methods scale with the number of neurons, inputs, outputs and dimensions in the NODE. The main challenge of these benchmarks is the presence of multiple NODEs within the GNODE. There are a total of 6 GNODEs (XS,S,M,L,XL, and XXL), all with the same number of layers. However, we increase the number of inputs, outputs, and parameters across all the layers of the GNODEs. Here, XS corresponds to the smallest model and XXL to the largest. A description of these architectures can be found in the Appendix.

Several trends can be observed from Table \ref{tab:randomRed}. The first is that in general, smaller models have smaller reach set computation times, with two notable exceptions: the first run with XS and the last experiment with XL. Furthermore, one can observe that the largest difference in the reach set computation times comes from increasing the number of states in the NODEs, which are $\{2,3,4,4,5,5\}$ for both NODEs in every model in $\{$XS,S,M,L,XL, and XXL$\}$ respectively, while increasing the input and output dimensions ($\{1,2,2,3,3,4\}$ respectively) does not affect the reachability computation as much. This is because of the complexity of nonlinear ODE reachability as state dimensions increase, while for $\mathscr{NN}-Layers$, increasing the size of the inputs or neurons by 1 or a few units does not affect the reachability computation as much.

\vspace{-1em}
\begin{table}[h!]
    \caption{Computation time of the reachability analysis of the randomly generated GNODEs. Results are shown in seconds.}
    \label{tab:randomRed}
    \centering
    \begin{adjustbox}{width=0.7\columnwidth,center}
    \addtolength{\tabcolsep}{8pt} 
    \begin{tabular}{l|cccccc}
    \centering 
     & \textbf{XS} & \textbf{S} & \textbf{M} & \textbf{L} & \textbf{XL} & \textbf{XXL} \\ 
    \hline 
    $\delta_{\mu}$ = 0.01 & 57.0 & 16.2 & 59.3 & 61.8 & 168.3 & 223.9 \\ 
    $\delta_{\mu}$ = 0.02 & 3.4 & 11.4 & 42.2 & 41.0 & 262.4 & 115.4 \\
    $\delta_{\mu}$ = 0.04 & 3.1 & 10.3 & 37.6 & 72.9 & 1226.3 & 243.6 \\ 
    \hline 
    \end{tabular}
    \addtolength{\tabcolsep}{-8pt} 
    \end{adjustbox}
\end{table}
\vspace{-1em}
\vspace{-1em}

\section{Related Work}



\textbf{Analysis of Neural ODEs.} To the best of our knowledge, this is the first empirical study of the formal verification of neural ODEs as presented in the general neural ODE (GNODE) class. Some other works have analyzed neural ODEs, but are limited to a more restricted class of neural ODEs with only purely continuous-time models. We refer to these models in this paper as NODEs. The most comparable work is a theoretical inquiry of the neural ODE verification problem using Stochastic Lagrangian Reachability (SLR) \cite{gruenbacher2021verification}, which was later extended and implemented in a stochastic reachability analysis tool called GoTube \cite{gruenbacher2021gotube}. The SLR method is an abstraction-based technique that is able to compute a tight over-approximation of the set of reachable states, and provide stochastic guarantees in the form of confidence intervals for the derived reachable set. 
Beyond reachability analysis, there have also been several works investigating the robustness of neural ODEs. In \cite{carrara2019wifs}, the robustness of neural ODE image classifiers is empirically evaluated against residual networks, which are a standard deep learning model for image classification tasks. Their analysis demonstrates that neural ODEs are more robust to input perturbations than residual networks. In a similar work, a robustness comparison of neural ODEs and standard CNNs was performed \cite{yan2020robustness}. Their work considers two standard adversarial attacks, Gaussian noise and FGSM \cite{goodfellow2015iclr}, and their analysis illustrate that neural ODEs are more robust to adversarial perturbations. 
In terms of the analysis of GNODEs, to the best of our knowledge, no comparable work as been done, although
for specific models where all the $\mathscr{NN}$-Layers of a GNODE have fully-connected layers with continuous differentiable activation functions like sigmoid or tanh, it may be possible to compare our methods to other tools (with minor modifications) like Verisig \cite{Ivanov2019verisig,ivanov2021verisig20} or JuliaReach \cite{bogomolov2019Juliareach}. However, that would restrict the more general class of neural ODEs (GNODEs) that we evaluate in this manuscript. 

\vspace{-1em}
\begin{table}[h!]
    \caption{Summary of related verification tools. A \checkmark means that the tool  supports verification of this class, a $\bigcirc$ means that it may be supported it, but some minor changes may be needed, a \textbf{--} means it does not support it, and a $\odot$ means that some small changes have been made to the tool for comparison and the tools has been used to verify at least one example on this class.}
    \label{tab:comparison}
    \begin{adjustbox}{width=0.86\columnwidth,center}
    \addtolength{\tabcolsep}{8pt} 
    \begin{tabular}{lccccc}
    \toprule
    \centering 
    \textbf{Tool} & \textbf{ODE}\footnote{} & \textbf{NODE}\footnote{} & \textbf{NN}\footnote{} & \textbf{NNCS} & \textbf{GNODE} \\ 
    \hline 
    CORA \cite{Althoff2015arch} & \checkmark & $\odot$ & -- & -- & -- \\
    ERAN \cite{singh2018neurips} & -- & -- & \checkmark & -- & -- \\
    Flow* \cite{Chen2013Flow*} & \checkmark & $\odot$ & -- & -- & -- \\
    GoTube \cite{gruenbacher2021gotube} & \checkmark & \checkmark & -- & -- & -- \\
    JuliaReach \cite{bogomolov2019Juliareach} & \checkmark & $\odot$ & \checkmark & \checkmark & -- \\
    Marabou \cite{katz2019marabou} & -- & -- & \checkmark & -- & -- \\
    nnenum \cite{bak2021nnenum} & -- & -- & \checkmark & -- & -- \\
    ReachNN \cite{Huang2019reachNN,fan2020reachNN*} & \checkmark & $\bigcirc$ & \checkmark & \checkmark & -- \\
    Reluplex \cite{katz2017reluplex} & -- & -- & \checkmark & -- & -- \\
    ReluVal \cite{wang2018reluval} & -- & -- & \checkmark & -- & -- \\
    Sherlock \cite{Dutta2019hscc} & \checkmark & $\bigcirc$ & \checkmark & \checkmark & -- \\
    SpaceEx \cite{Frehse2011spaceex} & \checkmark & $\bigcirc$ & -- & -- & -- \\
    Verisig \cite{Ivanov2019verisig,ivanov2021verisig20} & \checkmark & $\bigcirc$ & \checkmark & \checkmark & -- \\
    NNV \cite{tran2020cavtool} & \checkmark & $\odot$ & \checkmark & \checkmark & -- \\
    \textbf{NNVODE} (ours) & \checkmark & \checkmark & \checkmark & \checkmark & \checkmark \\
    \hline
    \end{tabular}
    \addtolength{\tabcolsep}{-8pt} 
    \end{adjustbox}
    \scriptsize{$^8$ODE verification is considered to be supported for any tool than can verify at least one of linear or nonlinear continuous-time ODEs. $^{9}$NODE is a specific type of ODE, so in theory any tool that supports ODE verification may be able to support NODE. However, these tools are optimized for NODE and in practice may not be able to verify most of these models as seen on our comparison with Flow*. $^{10}$We include in this category any tool that supports one or more verification methods for fully-connected, convolutional or pooling layers.}
\end{table}
\vspace{-1em}

\textbf{Verification of Neural Networks and Dynamical systems.}
The considered GNODEs are a combination of dynamical systems equations (ODEs) modeled using neural networks, and neural networks. When these subjects are treated in isolation, one finds that there are numerous studies that consider the verification of dynamical systems, and correspondingly there are numerous works that deal with the neural network verification problem. With respect to the former, the hybrid systems community has considered the verification of dynamical systems for decades and developed tools such as SpaceEx \cite{Frehse2011spaceex}, Flow* \cite{Chen2013Flow*}, CORA \cite{Althoff2015arch}, and JuliaReach \cite{bogomolov2019Juliareach} that deal with the reachability problem for discrete-time, continuous-time or even hybrid dynamics. A more comprehensive list of tools can be found in the following paper \cite{Doyen2018}. Within the realm of neural networks, the last several years have witnessed numerous promising verification methods proposed towards reasoning about the correctness of their behavior. Some representative tools include Reluplex \cite{katz2017reluplex}, Marabou \cite{katz2019marabou}, ReluVal \cite{wang2018reluval}, NNV \cite{tran2020cavtool} and ERAN \cite{singh2018neurips}. The tools have drawn inspiration from a wide range of techniques including optimization, search, and reachability analysis \cite{Liu2021}. A discussion of these approaches, along with pedagogical implementations of existing methods, can be found in the following paper \cite{Liu2021}.
Building on the advancements of these fields, as a natural progression, frameworks that consider the reachability problem involving neural networks and dynamical systems have also emerged. Problems within the space are typically referred to as Neural Network Control Systems (NNCS), and some representative tools include NNV \cite{tran2020cavtool}, Verisig \cite{Ivanov2019verisig,ivanov2021verisig20}, and ReachNN* \cite{Huang2019reachNN,fan2020reachNN*}. These tools have demonstrated their capabilities and efficiency in several works including the verification of an adaptive cruise control (ACC) \cite{tran2020cavtool}, an automated emergency breaking system \cite{tran2019emsoft}, and autonomous racing cars \cite{ivanov2020car,ivanov2021emsoft} among others, as well as participated in the yearly challenges of NNCS verification \cite{manzanas2019arch_ainncs,ARCH20AINNCS,ARCH2021ainncs}. These studies and their respective frameworks are very closely related to the work contained herein. In a sense, they deal with a restricted GNODE architecture, since their analysis combines the basic operations of neural network and dynamical system reachability in a feedback-loop manner. However, this restricted architecture would only consist of a fully-connected neural network followed by a NODE.
In Table \ref{tab:comparison}, we present a \textbf{comparison to verification tools} that can support one or more of the following verification problems: the analysis of continuous-time models with linear, nonlinear or hybrid dynamics (ODE), neural networks (NN), neural network control systems (NNCS), neural ODEs (NODE) and GNODEs as presented in Figure \ref{fig:gnode}. 

\section{Conclusion \texorpdfstring{$\&$}{} Future Work.} 
We have presented a verification framework to compute the reachable sets of a general class of neural ODEs (GNODE) that no other existing methods are able to solve. We have demonstrated through a comprehensive set of experiments the capabilities of our methods on dynamical systems, control systems and classification tasks, as well as the comparison to state-of-the-art reachability analysis tools for continuous-time dynamical systems (NODE). 
One of the main challenges we faced was the scalability of the nonlinear ODE reachability analysis as the dimension complexity of the models increased, as observed in the cartpole and damped oscillator examples. Possible improvements include integrating other methods into our framework such as \cite{li2020formats}, which improves the current nonlinear reachability analysis via an improved hybridization technique that reduces the sizes of the linearization domains, and therefore reduces overapproximation error. 
Another approach would be to make use of the Koopman Operator linearization prior to analysis in order to compute the reach sets of the linear system, which are easier and faster to compute as observed from the experiments conducted using the two-dimensional spiral benchmark \cite{Bak2021adhs}.
In terms of models and architectures, latent neural ODEs \cite{rubanova2019latent} and some of its proposed variations such as controlled neural DEs \cite{kidger2020neuralcde} and neural rough DEs \cite{morrill2021RDEs} have demonstrated great success in the area of time-series prediction, improving the performance of NODEs, ANODEs and other deep learning models such as RNNs or LSTMs in time-series tasks. The main idea behind these models is to learn a \emph{latent} space from the input of the neural ODE from which to sample and predict future values. In the future, we will analyze these models in detail and explore the addition of verification techniques that can formally analyze their behavior. 

\subsubsection*{Acknowledgements}
The material presented in this paper is based upon work supported by the National Science Foundation (NSF) through grant numbers 1910017 and 2028001, the Defense Advanced Research Projects Agency (DARPA) under contract number FA8750-18-C-0089, and the Air Force Office of Scientific Research (AFOSR) under contract number FA9550-22-1-0019. Any opinions, findings, and conclusions or recommendations expressed in this paper are those of the authors and do not necessarily reflect the views of AFOSR, DARPA, or NSF.
\small
\bibliographystyle{splncs04}
\bibliography{main.bib}




\appendix



\section{Appendix}

\subsection{Implementation Example of a NODE}
Let the function $g(z)$ be composed of two fully-connected layers, where $\sigma_1$ and $\sigma_2$ are \emph{tanh} and \emph{linear} activation functions, respectively. The state, $z$ $\in$ $\mathbb{R}^2$, is a two-dimensional vector, the first layer contains five neurons and the second layer has two neurons. Therefore, the parameters of the layers are $W_1 \in \mathbb{R}^{5 \times 2}$, $\textbf{b}_1$ $\in$ $\mathbb{R}^5$, $W_2$ $\in$ $\mathbb{R}^{2 \times 5}$ and $\textbf{b}_2$ $\in$ $\mathbb{R}^{2}$. We can define the state derivatives of this NODE as:
%
\begin{gather}
\label{eq:omegalayer_ex}
    \dot{z} = g(z) = L_2(L_1(z)) = \sigma_2(W_2 \times (\sigma_1(W_1 \times z + \textbf{b}_1) + \textbf{b}_2) \\  
    = W_2 \times (tanh(W_1 \times z + \textbf{b}_1) + \textbf{b}_2 . \nonumber
\end{gather}

For NNVODE as well as JuliaReach or GoTube, the representation in ~\eqref{eq:omegalayer_ex} is a valid format. However, for consistency and comparison purposes, other tools like Flow* require the explicit equation for every state $z_i$. Let the states $z$, weights $W_k$ and biases $\textbf{b}_k$ be

\begin{gather*}
z = \begin{bmatrix}
z_1 \\ z_2
\end{bmatrix},
W_1 = 
\begin{bmatrix} 
w_{1_{11}} & w_{1_{12}} \\
w_{1_{21}} & w_{1_{22}} \\
w_{1_{31}} & w_{1_{32}} \\
w_{1_{41}} & w_{1_{42}} \\
w_{1_{51}} & w_{1_{52}} \\ 
\end{bmatrix}, 
\textbf{b}_1 = 
\begin{bmatrix} 
b_{1_1} \\
b_{1_2}  \\
b_{1_3}  \\
b_{1_4}  \\
b_{1_5}  \\
\end{bmatrix}, \\
W_2 = 
\begin{bmatrix} 
w_{2_{11}} & w_{2_{12}} & w_{2_{13}} & w_{1_{14}} & w_{2_{15}} \\
w_{2_{21}} & w_{2_{22}} & w_{2_{23}} & w_{1_{24}} & w_{2_{25}} 
\end{bmatrix}, 
\textbf{b}_2 = 
\begin{bmatrix} 
b_{2_1} \\
b_{2_2} 
\end{bmatrix},
\end{gather*}
%
then we can explicitly write out the full equations for every state derivative:
%
\begin{gather*}
\dot{z}_1 = w_{2_{11}}  tanh(w_{1_{11}}z_1 + w_{1_{12}}z_2 + b_{1_1}) + w_{2_{12}} tanh(w_{1_{21}}z_1 \\
+ w_{1_{22}}z_2 + b_{1_2}) + w_{2_{13}} tanh(w_{1_{31}}z_1 +  w_{1_{32}}z_2 + b_{1_3}) \\
+ w_{1_{14}} tanh(w_{1_{41}}z_1 + w_{1_{42}}z_2 + b_{1_4}) \\ 
+ w_{2_{15}} tanh(w_{1_{51}}z_1 + w_{1_{52}} z_2 + b_{1_5}) + b_{2_1} , \\
\dot{z}_2 = w_{2_{21}}  tanh(w_{1_{11}}z_1 + w_{1_{12}}z_2 + b_{1_1}) +w_{2_{22}} tanh(w_{1_{21}}z_1 \\
+ w_{1_{22}}z_2 + b_{1_2}) + w_{2_{23}} tanh(w_{1_{31}}z_1 +  w_{1_{32}}z_2 + b_{1_3}) \\
+ w_{1_{24}} tanh(w_{1_{41}}z_1  + w_{1_{42}}z_2 + b_{1_4}) \\
+ w_{2_{25}} tanh(w_{1_{51}}z_1 + w_{1_{52}} z_2 + b_{1_5}) + b_{2_1} ,
\end{gather*}


For the special case where $\sigma_k$ is linear for every layer $k \in m$, we can simplify the NODE as a linear ODE:
%
\begin{equation}\label{eq:linearODE}
\begin{split}
    \dot{z} = & \left(W_m W_{m-1} \ldots W_1\right)z + \textbf{b}_m + \textbf{b}_{m-1}W_m \\ 
     & + \textbf{b}_{m-2}W_m W_{m-1} + \textbf{b}_1 W_m \ldots W_2 \\
    \dot{z} = & \left(W_m W_{m-1} \ldots W_1\right)z + \textbf{b}_1 W_m \ldots W_2 \\ 
    & + \textbf{b}_{2}W_m \ldots W_{3} + \textbf{b}_{m-1}W_m + \textbf{b}_m\\
%
%
    \dot{z} = & \prod_{i=1}^m (W_i) z + \sum_{i=1}^{m-1} (\textbf{b}_i \prod_{j=i+1}^m (W_j)) + \textbf{b}_{m}, \\
    \dot{z} = & Az + Bu + c, 
    \end{split}
\end{equation}

where $A$ is equal to weights term, $A$ = $\prod_{i=1}^m (W_i)$, $B$ = $0$, and the rest of the terms of the equation corresponds to the constant vector \\ $c$ = $\sum_{i=1}^{m-1}$ (\textbf{b}$_i$ $\prod_{j=i+1}^m (W_j))$ + \textbf{b}$_{m}$.

\subsection{Neural ODE architectures}

In this section we describe the architectures used for all the results explained in Section \ref{sec:eval}. We utilized the same base structure to explain all GNODEs, which consist on a set of $\mathscr{NN}$-Layers, then a NODE, and then another set of $\mathscr{NN}$-Layers, with the exception of the random experiments, which have a total of two NODEs. The names of the layers are abbreviated to include the full architecture in the tables, where layer ($ns_i$) corresponds to a weighted layer with activation function $layer$ and $ns_i$ neurons. If there is no number next to the layer it means that only the activation function is applied, no weights corresponding to that layer. When a set of layers are inside brackets, it means that this subset of layers is consecutively repeated $T$ times. For the description and architectures of the models of the fixed point attractor (FPA) and cartpole, we refer the reader to \cite{musau2018arch} and \cite{gruenbacher2020lagrangian} respectively.

\begin{table*}[]
\caption{\label{tab:NODEarchs} GNODE architectures of the spiral 2D benchmark, all the models of the damped oscillator (D.Osc.$_{n_{aug}}$), where n$_{aug} \in$ \{0,1,2\} corresponds to the number of augmented dimensions, and 6 classification models trained on MNIST.}
\centering
\begin{adjustbox}{width=\textwidth,center} 
\begin{tabular}{llll}
\multicolumn{1}{l}{\textbf{}} &
 \multicolumn{1}{l}{\textbf{$\mathscr{NN}$-Layers}} & \multicolumn{1}{l}{\textbf{NODE}} &  \multicolumn{1}{l}{\textbf{$\mathscr{NN}$-Layers}} \\ \hline
Spiral$_L$ & N/A & fc (10) - fc (2)  &  N/A  \\
Spiral$_NL$  & N/A  & tanh (10) - fc (2) & N/A  \\ 
D.Osc.$_{n_{aug}}$  & fc (2+n$_{aug}$)  & fc (20) - fc (20) -fc (2+n$_{aug}$) & fc(2)  \\
FNODE$_{S}$  & relu (64) - relu (10)  & fc (10)  & softmax (10)   \\
FNODE$_{M}$ & relu (64) - relu (32) - fc (16) & fc (10) - fc (16) & softmax (10) \\
FNODE$_{L}$  & relu (64) - [relu (32)] [x3] - fc (16) & [fc (10)] [x3] - fc (16) & softmax (10) \\
CNODE$_{S}$ & [conv2d (4,3) - BN - relu] [x2] - flatten & fc (10)  - fc (676) & softmax (10) \\
CNODE$_{M}$ & [conv2d (10,3) - BN - relu] [x2] - flatten & fc (10) - fc (1690) & softmax (10) \\ 
CNODE$_{L}$ & [conv2d (16,3) - BN - relu] [x2] - flatten & fc (10) - fc (2704) & softmax (10) \\
\bottomrule 
\end{tabular}
\end{adjustbox}
\end{table*}

\begin{table*}[]
\caption{\label{tab:randExarchs} GNODE architectures of the randomly generated GNODE with multiple NODEs. Next to the model, in parenthesis, is the input size of the GNODE. }
\centering
\begin{adjustbox}{width=\textwidth,center} 
\addtolength{\tabcolsep}{8pt} 
\begin{tabular}{l|lllll}
\multicolumn{1}{l}{\textbf{}} &
\multicolumn{1}{c}{\textbf{$\mathscr{NN}$-Layer}} & \multicolumn{1}{c}{\textbf{NODE}} &  \multicolumn{1}{c}{\textbf{$\mathscr{NN}$-Layer}} & 
\multicolumn{1}{c}{\textbf{NODE}} &  \multicolumn{1}{c}{\textbf{$\mathscr{NN}$-Layer}}\\ \hline
XS (1)  & tanh(2) & tanh(2) - tanh(2) & tanh(2) & tanh(2) & tanh(1)   \\
S (2)   & tanh(3) & tanh(5) - tanh(3) & tanh(3) & tanh(3) & tanh(2)   \\
M (2)   & tanh(4) & tanh(8) - tanh(4) & tanh(4) & tanh(4) & tanh(2)   \\
L (3)   & tanh(4) & tanh(8) - tanh(4) & tanh(4) & tanh(4) & tanh(3)   \\
XL (3)  & tanh(5) & tanh(10) - tanh(5) & tanh(5) & tanh(5) & tanh(3)  \\
XXL (4) & tanh(5) & tanh(10) - tanh(5) & tanh(5) & tanh(5) & tanh(4) \\
\hline
\end{tabular}
\addtolength{\tabcolsep}{-8pt} 
\end{adjustbox}
\end{table*}

\noindent\textbf{ACC.} The ACC GNODEs used as plant models for the NNCS reachability are represented as a 3$^{rd}$ order NODE, building upon prior knowledge of the ego and lead car dynamics. The dynamics are defined in ~\eqref{eq:accDyns}, and the NODE architectures are described in Table \ref{tab:accarchs}.

\begin{equation}\label{eq:accDyns}
    \begin{split}
    \dot{x}_1 & = \dot{x}_{\textrm{lead}} = v_{\textrm{lead}}, \\ 
    \dot{x}_2 & =\dot{v}_{\textrm{lead}} = \gamma_{\textrm{lead}}, \\ 
    \dot{x}_3 & = \dot{\gamma}_{\textrm{lead}} = \dot{y}_1, \\ 
    \dot{x}_4&  = \dot{x}_{\textrm{ego}} = v_{\textrm{ego}}, \\ 
    \dot{x}_5 & = \dot{v}_{\textrm{ego}} = \gamma_{\textrm{ego}}, \\ 
    \dot{x}_6 & = \dot{\gamma}_{\textrm{ego}} = \dot{y}_2, \\
    \dot{y} & = g_{acc}(z_{acc}),  \\
    z_{acc} & = [\gamma_{\textrm{lead}},\gamma_{\textrm{ego}}, a_{\textrm{lead}}, a_{\textrm{ego}}]^{\intercal}
    \end{split}
\end{equation}
where $acc$ = $\{$L, NL$\}$, corresponding to the linear and nonlinear NODEs. 

\begin{table*}[]
\caption{\label{tab:accarchs} Architectures of the ACC 3$^{rd}$ order NODEs, where $g_{_L}$ represents the NODE architecture of the linear model and $g_{_{NL}}$ the nonlinear one. }
\centering
\begin{adjustbox}{width=0.45\textwidth,center} 
\addtolength{\tabcolsep}{8pt} 
\begin{tabular}{l|c}
\multicolumn{1}{l}{\textbf{}} & \multicolumn{1}{c}{\textbf{NODE}}  \\ \hline
$g_{_NL}$ & tanh(10) - tanh(4) \\
$g_{_{L}}$ & fc(20) - fc(4) \\
\hline
\end{tabular}
\addtolength{\tabcolsep}{-8pt} 
\end{adjustbox}
\end{table*}

\subsection{Evaluation - Reachability Analysis Parameters}
In this section we describe the parameters used for each of the tools we compare against, as well as the the parameters used in NNVODE. For each benchmark, we create an ordered list of parameters based on the order as these are presented in the corresponding results table. If there is only one value for the parameters, it means that all the reachability parameters are kept constant throughout the benchmark experiments.

\subsection*{Spiral2D}

\textbf{Flow*}
\begin{itemize}
    \item Reach step: $\{0.01, 0.01, 0.01, 0.002, 0.002, 0.002\}$
    \item Taylor Models: $\{8, 8, 8, [6,20], [6,20], [6,20], \}$
\end{itemize}

\hspace{-1em}\textbf{GoTube}
\begin{itemize}
    \item Reach step: 0.01
    \item Batch size: 1000
    \item gamma: 0.01
    \item mu: 1.5
\end{itemize}

\hspace{-1em}\textbf{JuliaReach}
\begin{itemize}
    \item alg: TMJets21a
    \item abstol: 1$e^{-10}$
    \item orderT: 5
    \item orderQ: 1
    \item Max steps: 3500
    \item No parameters were specified for the linear model (LinearTS), we run all instances with the default parameters using $solve(problem, t_F)$
\end{itemize}


\subsection*{Fixed Point Attractor (FPA)}

\textbf{Flow*}
\begin{itemize}
    \item Reach step:  0.01
    \item Taylor Models: $\{8, 20, 30\}$
\end{itemize}

\hspace{-1em}\textbf{GoTube}
\begin{itemize}
    \item Reach step: 0.01
    \item Batch size: 1000
    \item gamma: 0.01
    \item mu: 1.5
\end{itemize}

\hspace{-1em}\textbf{JuliaReach}
\begin{itemize}
    \item alg: TMJets21a
    \item abstol: 1$e^{-10}$
    \item orderT: 5
    \item orderQ: 1
    \item Max steps: $\{ 400, 1700, 3500 \}$
\end{itemize}


\subsection*{Cartpole}

\textbf{Flow*}
\begin{itemize}
    \item Time step: 0.01
    \item Taylor models: $\{30, 20, 8\}$
\end{itemize}

\hspace{-1em}\textbf{GoTube}
\begin{itemize}
    \item Reach step: 0.01
    \item Batch size: 1000
    \item gamma: 0.01
    \item mu: 1.5
\end{itemize}

\hspace{-1em}\textbf{JuliaReach}
\begin{itemize}
    \item alg: TMJets21a
    \item abstol: 1$e^{-10}$
    \item orderT: 5
    \item orderQ: 1
    \item Max steps: $\{ 400, 1700, 3500 \}$
\end{itemize}

\end{document}